\documentclass{article}

\usepackage{PRIMEarxiv}

\usepackage[utf8]{inputenc} 
\usepackage[T1]{fontenc}    
\usepackage{hyperref}       
\usepackage{url}            
\usepackage{booktabs}       
\usepackage{amsfonts}       
\usepackage{nicefrac}       
\usepackage{microtype}      
\usepackage{lipsum}
\usepackage{fancyhdr}       
\usepackage{graphicx}       
\graphicspath{{media/}}     

\usepackage{multirow}
\usepackage{makecell}
\usepackage{subfig}

\title{A multi-scale vision transformer-based multimodal GeoAI model for mapping Arctic permafrost thaw
}

\author{
  Wenwen Li, Chia-Yu Hsu, Sizhe Wang, Zhining Gu \\
  School of Geographical Sciences and Urban Planning \\
  Arizona State University \\
  Tempe, AZ, USA\\
  \texttt{\{wenwen, chsu53, wsizhe, zhiningg\}@asu.edu} \\
   \And
  Yili Yang, Brendan M. Rogers, Anna Liljedahl \\
  Woodwell Climate Research Center \\
  Falmouth, MA, USA \\
  \texttt{\{yyang, brogers, aliljedahl\}@woodwellclimate.org} \\
}

\begin{document}
\maketitle

\begin{abstract}
Retrogressive Thaw Slumps (RTS) in Arctic regions are distinct permafrost landforms with significant environmental impacts. Mapping these RTS is crucial because their appearance serves as a clear indication of permafrost thaw. However, their small scale compared to other landform features, vague boundaries, and spatiotemporal variation pose significant challenges for accurate detection. In this paper, we employed a state-of-the-art deep learning model, the Cascade Mask R-CNN with a multi-scale vision transformer-based backbone, to delineate RTS features across the Arctic. Two new strategies were introduced to optimize multimodal learning and enhance the model’s predictive performance: (1) a feature-level, residual cross-modality attention fusion strategy, which effectively integrates feature maps from multiple modalities to capture complementary information and improve the model's ability to understand complex patterns and relationships within the data; (2) pre-trained unimodal learning followed by multimodal fine-tuning to alleviate high computing demand while achieving strong model performance. Experimental results demonstrated that our approach outperformed existing models adopting data-level fusion, feature-level convolutional fusion, and various attention fusion strategies, providing valuable insights into the efficient utilization of multimodal data for RTS mapping. This research contributes to our understanding of permafrost landforms and their environmental implications.
\end{abstract}


\section{Introduction}
As awareness grows, it becomes evident that the temperature in Arctic regions has been rising at a rate at least twice as fast as the global average increase \cite{chylek2022arctic}. Permafrost landscapes are significantly impacted by Arctic warming \cite{jones2019rapid}. Abrupt permafrost thaw and climate shifts (e.g., increasing temperature and precipitation) have caused active layer detachment or slope failure, expediting the formation of distinctive landforms known as Retrogressive Thaw Slumps (RTSs) \cite{nitze2021deep, huang2022accuracy, turner2021detailed}. The progression of RTSs exposes underlying soil that contains abundant organic materials, increasing the risk of accumulation of heavy metal chemicals \cite{huang2020using, huang2022accuracy}. Permafrost thaw is damaging vegetation cover and disrupting hydrological and ecological processes in surrounding areas \cite{witharana2022automated}. Both the ecological structure and infrastructure are vulnerable to the impacts of permafrost thaw. Additionally, the intensification of carbon emissions is another rising concern associated with this process \cite{witharana2022automated}. As a result, understanding RTS dynamics becomes pivotal for understanding permafrost thaw and its impacts on terrestrial alterations, local ecosystems, the hydrological system, and the carbon cycle \cite{witharana2022automated, yang2023mapping}.

Mapping RTSs offers insights into the spatiotemporal changes in the permafrost landscape and local environmental dynamics. However, no pan-Arctic maps of RTS exist, hindering our ability to assess their full distribution, changes over time, and impacts on ecosystems and the carbon cycle. Traditionally, RTS mapping relied heavily on manual digitization using the Esri ArcGIS platform \cite{lantuit2008fifty, ramage2017terrain}. However, this method is very time-consuming, and is limited to small-area research \cite{huang2022accuracy, yang2023mapping}. Recent advancements in Geospatial Artificial Intelligence (GeoAI) \cite{li2020geoai} and Deep Learning (DL) have led to new models demonstrating outstanding performance in various computer vision tasks, including object detection \cite{cai2019cascade, feng2020deep, huang2020multi, hsu2021knowledge, bayoudh2022survey, li2022geoai}, and semantic and instance segmentation \cite{he2017mask, cai2019cascade, li2022real}. Specifically, semantic segmentation classifies every pixel in an image into predefined categories without differentiating individual objects of the same class. Object detection, on the other hand, identifies and localizes objects by predicting bounding boxes around them along with their associated class labels. Instance segmentation combines these goals by classifying each pixel while distinguishing between distinct instances of the same object class.

These cutting-edge GeoAI techniques have been employed to support diverse geographical applications, including plane and airport detection from aerial images \cite{xia2018dota}, land cover segmentation \cite{ulmas2020segmentation}, flood mapping \cite{lee2024improving}, and crop yield estimation \cite{xu2021interpreting}. In recent years, advanced GeoAI models have started to be applied to enhance the automation capabilities in RTS mapping. Semantic segmentation models, such as U-Net \cite{ronneberger2015unet}, DeepLabV3 \cite{chen2017rethinking}, and DeepLabV3+ \cite{chen2018encoder}, have been frequently employed on remote sensing images to delineate RTSs in the Arctic \cite{nitze2021deep, yang2023mapping} and on the Tibetan Plateau \cite{huang2020using}, providing valuable insights into the potential of DL models for RTS research. 

While the pioneering research mentioned above has significantly advanced the automated mapping of RTSs, prior studies have typically formulated RTS segmentation as a semantic segmentation problem. Although these methods can identify RTS and non-RTS pixels, they cannot distinguish RTSs as individual objects. Due to spatiotemporal variations in soil moisture, vegetation cover, water content, and terrain, the shapes of RTSs and their rates of change can vary \cite{hayes2022massive, yang2023mapping}. Additionally, each RTS uniquely impacts its local environment \cite{hayes2022massive}. Therefore, providing fine-grained, instance-level segmentation of RTSs is crucial for tracking their spatial extent, growth patterns, and characteristics over time, thereby enhancing the understanding of permafrost dynamics across heterogeneous Arctic landscapes.

Instance segmentation effectively identifies different objects from the image, which is suitable for recognizing individual RTS features. The process can employ GeoAI and deep learning models like Convolutional Neural Networks (CNNs), which utilize CNN-based encoders such as ResNet \cite{he2015delving} and VGG-Net \cite{simonyan2015vgg}. An alternative approach involves Vision Transformer (ViT)-based feature extractors. Originally developed for natural language processing, Transformers have been adapted for visual tasks through the development of ViT, enhancing their applicability in image segmentation tasks. Various instance segmentation models, adopting either a transformer or a CNN backbone, typically use an encoder followed by a decoder, also known as a segmentation head. Popular models include Mask R-CNN \cite{he2017mask} and Cascade Mask R-CNN \cite{cai2019cascade}. Mask R-CNN is a two-stage model which first generates proposals for object regions and then, in the second stage, refines these proposals by classification and regression tasks. Cascade Mask R-CNN makes further improvements to incorporate multiple stages to refine predictions in every stage, which achieves better performance than Mask R-CNN \cite{cai2019cascade}.

In addition, multimodal learning has further empowered vision tasks \cite{bayoudh2022survey, zhang2021deep}. “Multimodal” refers to the incorporation of various types of data inputs, such as images and text, to enrich the learning process. Each modality contributes unique and complementary information \cite{xu2023multimodal}, allowing for the generation of important features related to the targets. As a result, the integration of multiple modalities holds the potential to provide comprehensive feature information, serving as prior knowledge for GeoAI models to acquire rich information through joint learning \cite{zhang2021deep}. In the geospatial domain, the integration of different data modalities from different sources provides multifaceted characteristics of landforms. For instance, optical images visually distinguish landforms such as valleys, hills, and rivers. Multi-spectral images record data within specific wavelength bands beyond what is visible to human eyes, enabling in-depth analysis in various fields like agriculture and environmental monitoring. Light Detection and Ranging (LiDAR) uses laser light to measure distances, thereby generating high-resolution elevation models that offer detailed insights into the topography of landforms. 

This paper aims to take advantage of multimodal data across the Arctic and cutting-edge deep learning models to achieve automated RTS delineation. This is achieved by the introduction of a new multimodal learning strategy and the utilization of vision-transformer-based multi-scale instance segmentation techniques. The rest of the paper will be organized as follows. Section 2 will introduce relevant work on GeoAI and deep learning models and multimodal data usage in RTS mapping. Section 3 describes study areas for RTS mapping. Section 4 describes our proposed methodology in detail. Section 5 presents the experimental results and analysis. Section 6 concludes our work and proposes future research directions.

\section{Related Work}
\subsection{Deep learning-based RTS mapping}
Recent years have witnessed exciting advances in artificial intelligence techniques, particularly in object detection and segmentation tasks within computer vision. The use of segmentation DL models for RTS mapping is becoming a popular and promising solution for large-scale studies \cite{huang2020using, nitze2021deep, huang2022accuracy, udawalpola2022automated, witharana2022automated, yang2023mapping}. These models have been employed to assess RTS mapping performance, including U-Net \cite{nitze2021deep, witharana2022automated, yang2023mapping}, U-Net++ \cite{nitze2021deep}, DeepLabv3 \cite{nitze2021deep}, and DeepLabv3+ \cite{huang2020using, huang2022accuracy}. For instance, \cite{nitze2021deep} used DeepLabv3 and U-Net++ to map RTS and achieved an Intersection over Union (IoU) score of 0.58 with U-Net++. \cite{yang2023mapping} expanded the spatial extent of the RTS dataset and enhanced the DEM data by employing relative elevation and enhanced shaded-relief layer. By applying a label sampling approach and label smoothing strategy, the model achieved the highest IoU score of 0.71 with U-Net3+, outperforming the U-Net++, TransU-Net, and ResU-Net models. 

All these studies treat RTS delineation as a semantic segmentation task. However, as \cite{ramage2017terrain} pointed out, RTS landforms are dynamic, under varying environmental conditions including precipitation, temperature, and stream erosion influencing spatial and temporal variations in physical properties such as soil moisture, vegetation cover, and elevation contrast \cite{yang2023mapping}. For instance, while RTSs in Banks Island are mainly found along lake shores and valley slopes \cite{nitze2021deep}, those near the Lena River are largely characterized by dense shrubs along lake shores \cite{nitze2021deep}. Despite these variations, current studies have not focused on mapping and analyzing the unique characteristics of individual RTS features. Fortunately, instance segmentation models offer a solution to address this gap. 

Instance segmentation models, such as Mask R-CNN, incorporate Regional Proposal Network (RPN) to generate candidate object proposals based on generated feature maps and achieve the parallel classification, bounding box regression, and mask generation. Using this approach, individual objects, such as RTS, can be segmented. In our paper, we employed an enhanced model of Mask R-CNN, integrating it with a multi-scale vision transformer-based backbone and a multi-stage segmentation to further improve its segmentation performance.

\subsection{Multimodal learning in landform mapping}
In natural feature detection, the use of multi-modalities ranging from optical image and other data modalities has grown \cite{bayoudh2022survey, zhang2021deep, wang2021geoai}. By fusing heterogeneous data sources, models harness complementary features through joint learning, leading to optimized prediction outcomes \cite{zhang2021deep}. To effectively delineate RTS, \cite{huang2020using} combined optical images with DEM and its derivatives (e.g., slope and topographic position index), while \cite{yang2023mapping} incorporated NDVI and other DEM features including relative elevation and enhanced shaded-relief features. While these studies successfully leverage multimodal data to map landform features, the fusion strategies employed are relatively traditional, and a gap remains in integrating cutting-edge multimodal fusion strategies into landform mapping research.

For example, many studies in RTS mapping \cite{huang2020using, nitze2021deep, witharana2022automated, yang2023mapping} have employed a data-level fusion approach, in which additional data modalities are integrated as extra channels alongside RGB channels to serve as inputs to DL models. For instance, \cite{yang2023mapping} adapted the U-Net model, which originally takes 3 channels (red, green, and blue) as inputs. They expanded the input by concatenating the RGB image with NDVI, relative elevation, and enhanced shaded relief, resulting in a hybrid 6-channel input.

An alternative approach to multimodal fusion is feature-level fusion. In this method, feature maps of each modality are extracted independently by backbone feature extractors, such as ResNet \cite{he2015delving}, VGG-Net \cite{simonyan2015vgg}, and Multi-scale ViT \cite{li2022mvitv2}. To effectively integrate diverse types of information, different fusion strategies can be employed on multimodal feature maps. For instance, in mapping natural features, \cite{wang2021geoai} employed CNN-based backbones to individually extract feature maps for each modality, such as optical image, DEM, and DEM derivatives. They then introduced an additional convolutional fusion layer to integrate these modalities. Similarly, \cite{ophoff2019exploring} leveraged the RGB image and depth data to extract relevant features. After concatenating these features, they added a convolutional layer to merge them, enhancing the model's capability for subsequent object detection tasks involving classification and regression.

In addition to convolutional fusion, attention mechanisms are a powerful tool in deep learning. Attention modules can dynamically focus on the most informative parts (e.g., where the “attention” should be) of input data during processing \cite{vaswani2017attention}. By assigning varying weights to different elements, attention enables deep learning models to capture complex relationships and dependencies, even across distant features. For images, attention operates by dividing an image into patches or spatial regions, computing attention scores to identify which regions are most relevant for a given task \cite{dosovitskiy2021image, liu2021swin}. These scores weigh the contributions of each region, allowing the model to prioritize important features while de-emphasizing less informative areas. Extending the attention mechanism in a multimodal learning framework can further enhance the “attention” capturing capabilities by leveraging relevant information from different modalities to reinforce the feature extraction process. This approach reinforces important cross-modal features and creates a more informative feature representation, resulting in stronger model prediction capabilities. 

Despite its promise, comprehensive research on multimodal fusion strategies regarding their effectiveness, computational efficiency, contributions of data modalities, and the methods’ generalizability has been underexplored in landform mapping applications, especially for mapping retrogressive thaw slumps. To address this gap, our research has proposed a novel residual cross-modality fusion strategy that effectively integrates multimodal data with varying characteristics to advance RTS mapping across the heterogeneous Arctic landscape. The next section introduces the study area, followed by a detailed description of our proposed model in Section 4.

\begin{figure}[!t]
\centering
\includegraphics[width=.95\textwidth]{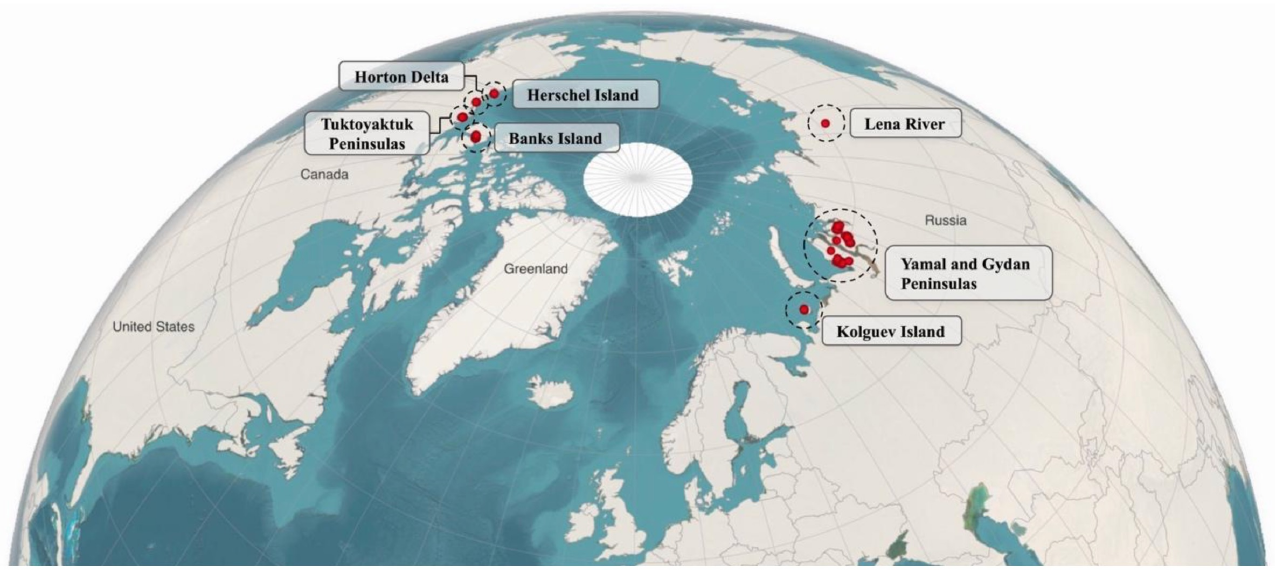}
\caption{Study Areas. There are 855 RTS samples in total located in seven sites from both Canada and Russia. In Canada, study sites include Herschel Island (41 RTS samples), Horton Delta (43 RTS samples), Tuktoyaktuk peninsulas (165 RTS samples), and Banks Island (174 RTS samples). In Russia, 399 RTSs are from Yamal and Gydan peninsulas, 7 samples are near the Lena River, and the remaining 26 RTS samples are from Kolguev Island.}
\label{fig_study_area}
\end{figure}
\section{Study Area}

Our study area encompasses seven sites, utilizing data from \cite{yang2023mapping} and \cite{nitze2021deep}. These sites provide a diverse representation of RTSs, covering various environmental and geomorphological conditions. Three sites, including the Yamal and Gydan Peninsulas, the Lena River, and Kolguev Island, are located in Russia. The other four, situated in Canada, comprise Herschel Island, Horton Delta, the Tuktoyaktuk Peninsula, and Banks Island. Figure \ref{fig_study_area} illustrates the spatial distribution of the RTSs. For the experiment, we collected a total of 855 RTS image scenes, with 399 samples from the Yamal and Gydan Peninsulas, 7 near the Lena River, and 26 from Kolguev Island in Russia. Additionally, 41 samples are from Herschel Island, 43 from Horton Delta, 165 from the Tuktoyaktuk Peninsula, and 174 from Banks Island in Canada.

Each study site presents distinctive environmental conditions for RTSs. For instance, the Yamal and Gydan peninsulas are the only known areas with gas emission craters \cite{zolkos2021detecting}. In addition, Banks Island and Herschel Island are characterized by extensive ground ice and exhibit the most active RTSs \cite{nitze2021deep}. Herschel Island is further distinguished by shrubby tundra interspersed with small lakes and streams. Horton Delta, in contrast, features steep terrain dominated by the vegetation of dwarf shrub tundra, where RTSs are commonly found on steep slopes. Moreover, Kolguev Island, with its ice-rich permafrost, often sees RTSs on steep coastal bluffs \cite{nitze2021deep}.

\section{Methodology}
\begin{figure*}[!t]
\centering
\includegraphics[width=.95\textwidth]{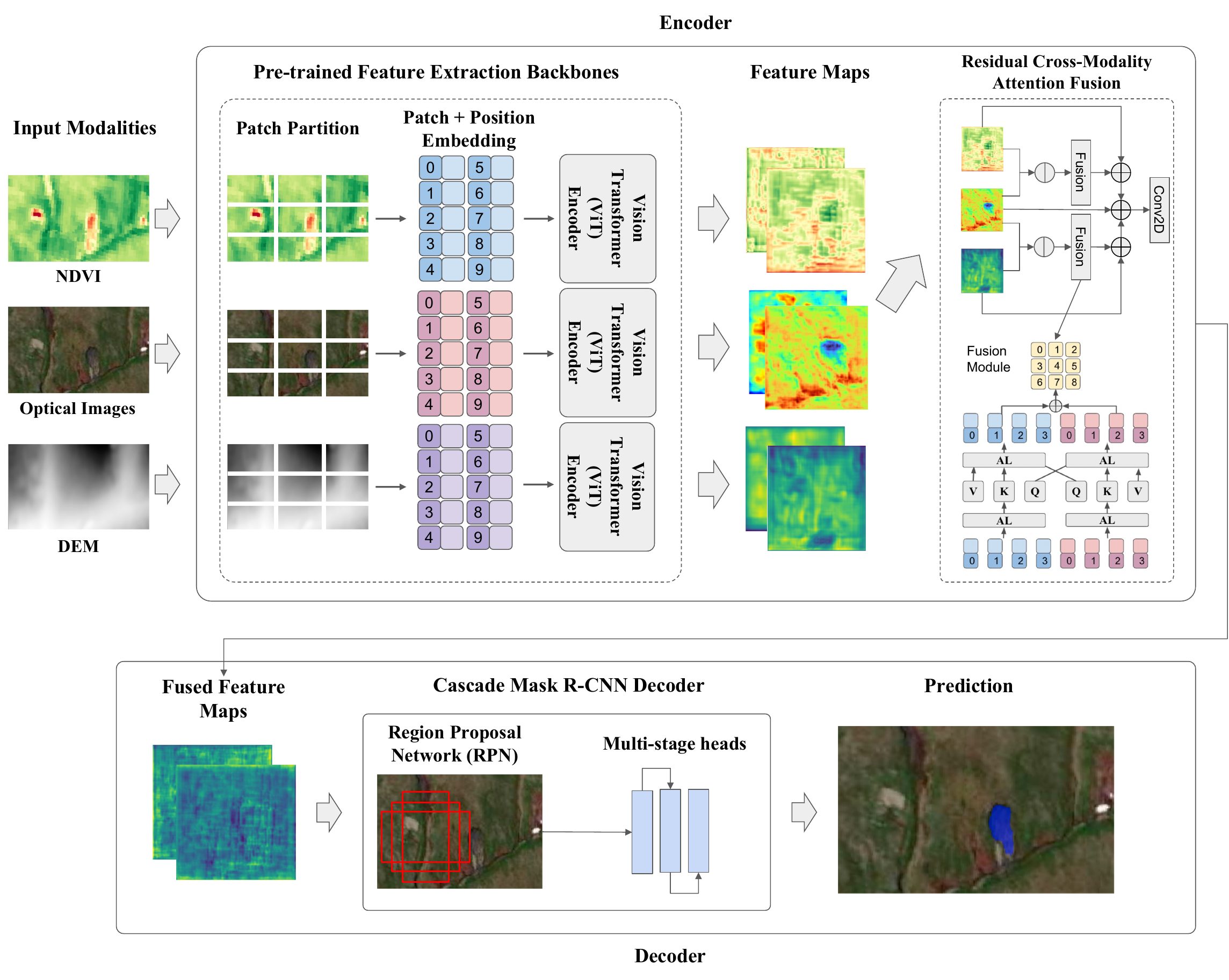}
\caption{Architecture of multimodal instance segmentation model for RTS mapping.}
\label{fig_arch}
\end{figure*}

To segment RTS features with high accuracy, we implemented our multimodal fusion framework by extending the Cascade Mask R-CNN instance segmentation model \cite{cai2019cascade}. Such an instance segmentation pipeline (Figure \ref{fig_arch}) starts with the backbone network (the encoder) to extract feature maps from input images as their encoded feature representation. We chose a multi-scale ViT as the model backbone because of its ability to capture global contextual information at varying scales and resolutions. The decoder of the instance segmentation model utilizes the resultant feature maps to reconstruct spatial details for accurate instance segmentation.

\begin{figure}[!b]
\subfloat[RGB]{\includegraphics[width=.3\textwidth]{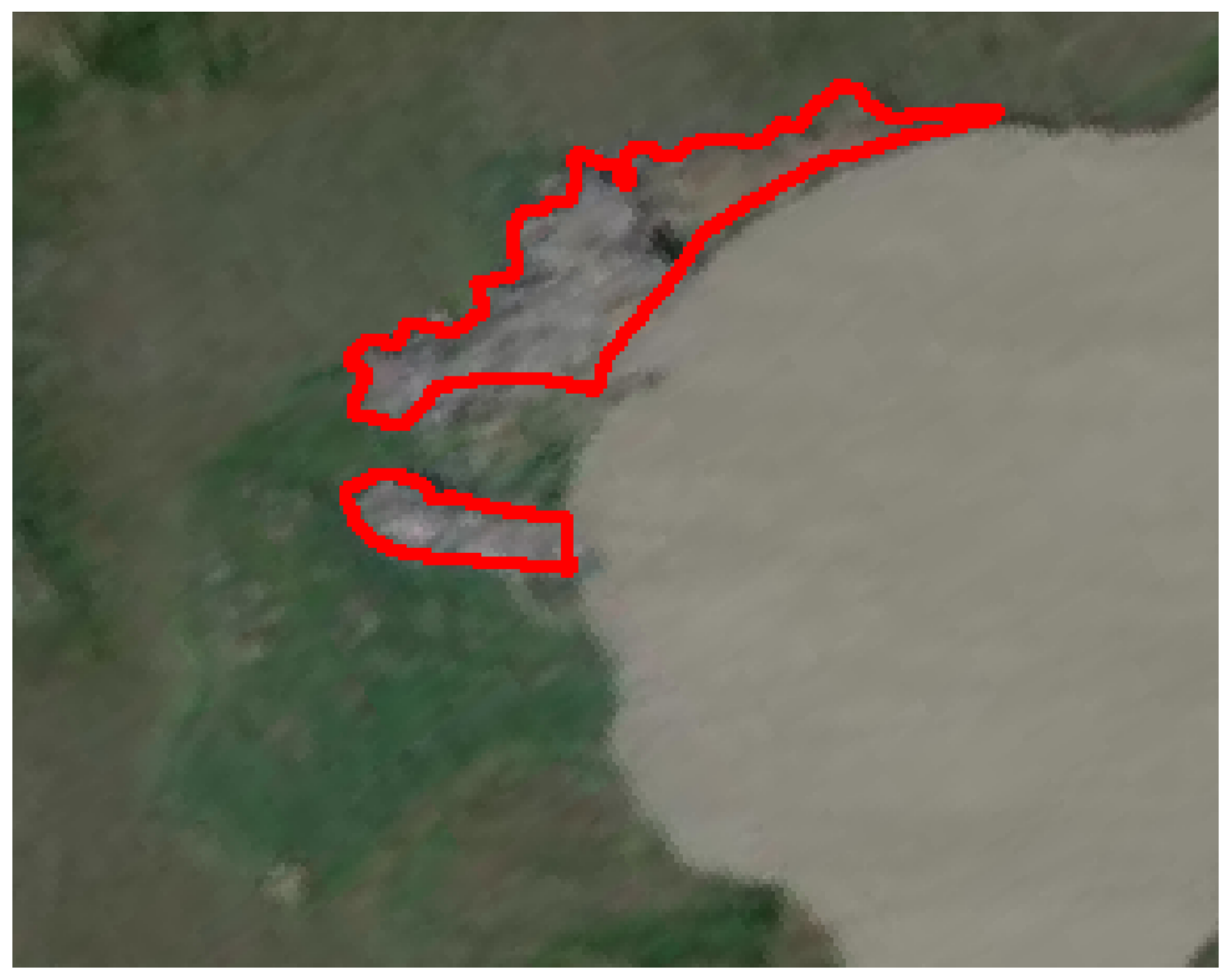}%
\label{fig_exp_data_rgb}}
\hfill
\subfloat[NDVI]{\includegraphics[width=.3\textwidth]{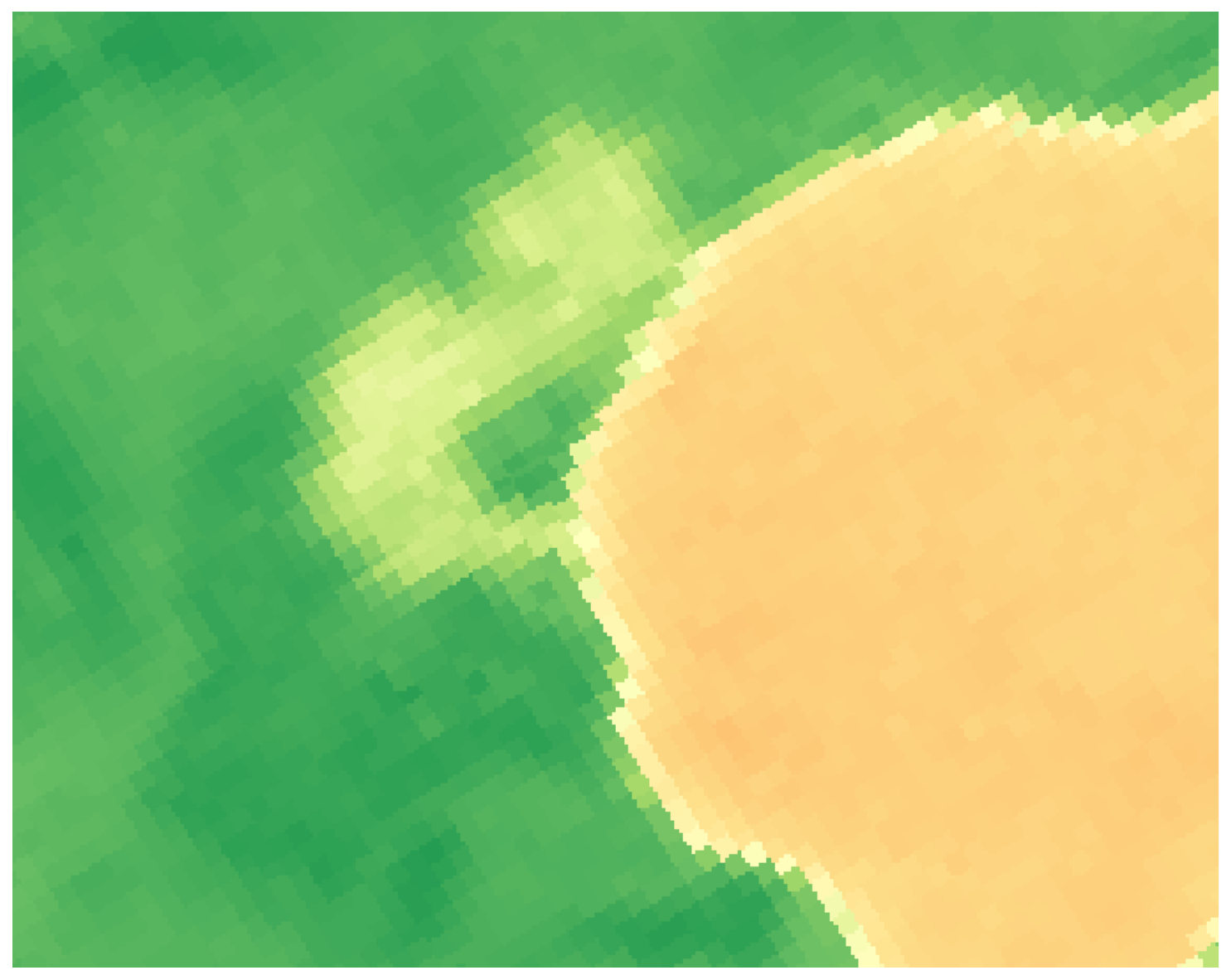}%
\label{fig_exp_data_ndvi}}
\hfill
\subfloat[NIR]{\includegraphics[width=.3\textwidth]{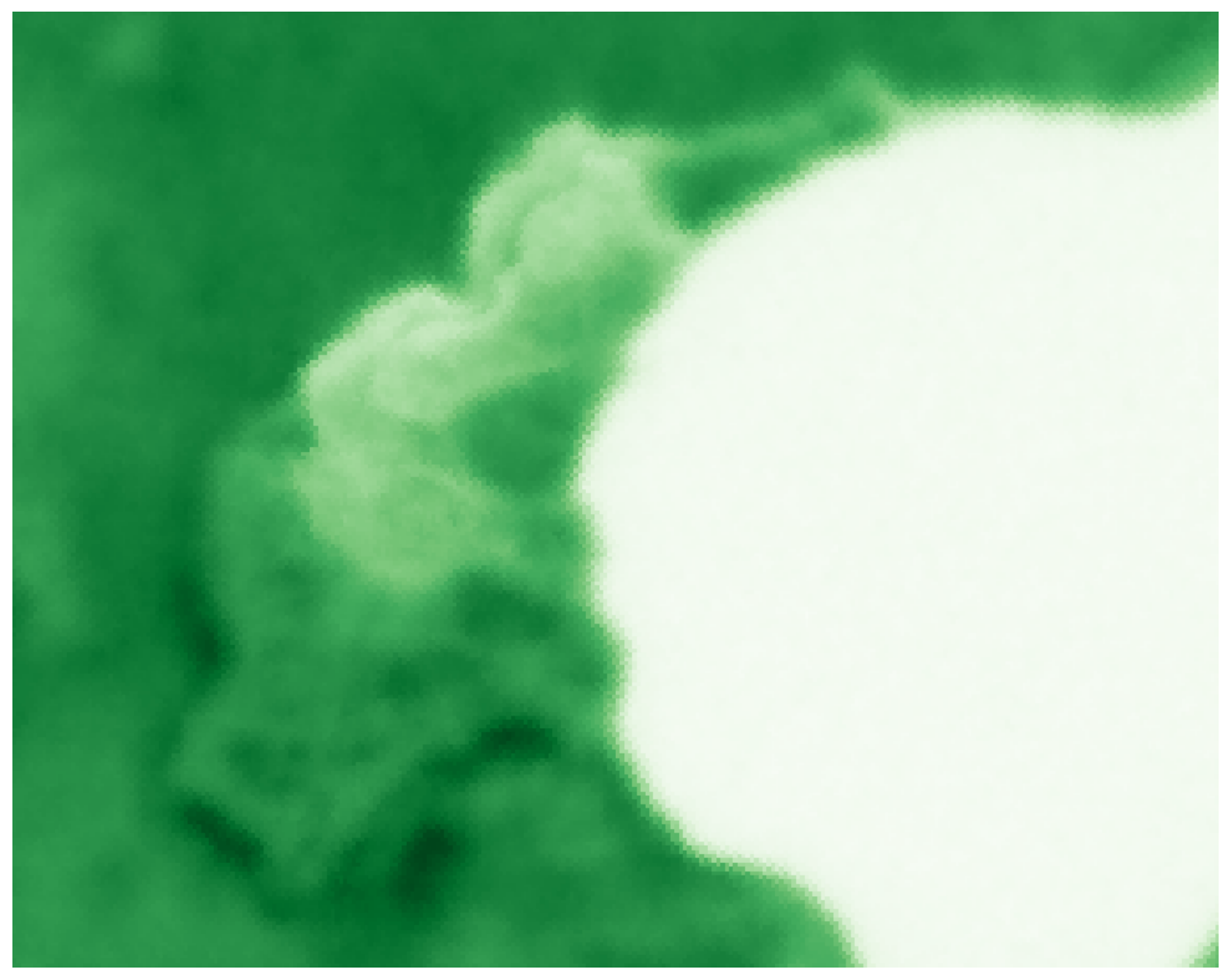}%
\label{fig_exp_data_nir}}
\caption{Example multimodal training data and labels (in red) for RTS segmentation.}
\label{fig_exp_data}
\end{figure}

The aforementioned model was designed to use a single modality as input. As mentioned before, multimodal data provide complementary information to support a better understanding and delineation of RTS features. For example, optical images with RGB bands support the visual inspection of RTS. Since RTS results from landslides and ground subsidence, this process often causes a change in vegetation cover around it. As such, vegetation indices (e.g., NDVI) and spectral bands beyond the visible spectrum, such as Near Infrared (NIR), become critical additional data modalities for capturing vegetation changes in RTS regions relative to their surroundings. Figure \ref{fig_exp_data} illustrates an example of the multimodal data and the corresponding RTS labels created as part of the AI-ready training dataset.

To better accommodate multiple data modalities, we propose two innovative strategies for the image segmentation pipeline: (1) a feature-level residual cross-modality attention fusion and (2) pre-trained unimodal learning followed by multimodal model fine-tuning. Both strategies, as illustrated in Figure \ref{fig_arch}, focus on modifying the instance segmentation pipeline to enable effective multimodal learning. During pre-trained unimodal learning, each input modality (RGB, NDVI, or NIR) is individually trained on a Cascade Mask R-CNN model with a multi-scale ViT encoder, generating backbone weights optimized for segmentation accuracy. These pre-trained backbone models are then utilized in the multimodal training process.
With feature maps for three modalities extracted, the residual cross-modality attention fusion strategy (Figure \ref{fig_arch}, top right) is applied to generate fused feature maps by combining complementary information across modalities. The resultant feature maps are then sent to the decoder, composed of a Region Proposal Network and multi-stage segmentation heads for instance classification, detection refinement, and final segmentation.

\subsection{Multi-scale Vision-Transformer-based feature extraction backbone}
In this work, we adopt a multi-scale vision transformer as the backbone model at the pre-training phase. The backbone, also referred to as the encoder, transforms the input image into feature maps. These feature maps are then processed by the segmentation head to reconstruct images for the final predictions in what is also known as the decoding stage.
Different from CNN-based backbones which take the entire image as input, ViT divides each image into uniform-sized patches and takes each flattened image patch as input. These flattened patches, each augmented with position embeddings, are then fed into a Transformer Encoder. Multi-scale ViT builds upon the standard ViT by transitioning from a constant resolution to multi-scale resolutions. This is achieved by increasing the number of channels and decreasing the sequence length (or resolution) at various scaled stages \cite{fan2021multiscale}. The integration of relative position embeddings and the enhanced pooling connection within the Transformer Encoder significantly improves the model’s performance. These enhancements allow the model to consider image semantic information and spatial relationships based on relative locations, which helps reduce computational and memory demands.

\subsection{Singular-modal pretraining and multimodal fine-tuning}
A challenge for multimodal learning, especially at the feature fusion level, is the memory consumption required for multi-backbones, each dedicated to supporting feature extraction from a single modality. Transformer architectures also often require significantly more memory than traditional CNN-based models due to the complex attention mechanisms they adopt. As a result, introducing multiple feature extraction paths (as shown in Figure \ref{fig_arch}) typically demands substantial computing resources and GPU capacity, making model training more challenging than in unimodal learning. Furthermore, larger models tend to be more difficult to train and converge. When training data is limited, the model parameters may not be optimally tuned and, therefore, may fail to achieve optimal performance.

To address this challenge, we proposed a phased training strategy, beginning with pre-trained single-modal learning, followed by multimodal model fine-tuning. Each modality's data is first processed through a unimodal instance segmentation pipeline for model pre-training. The backbone weights yielding the best model results for each modality are saved in preparation for subsequent phases of multimodal fine-tuning. Once the best model prediction accuracy is achieved for each modality, the corresponding ViT backbone weights are frozen, making them untrainable. During the multimodal learning stage, each data modality passes through its pre-trained multiscale ViT backbone to extract the most effective feature maps, which are then utilized for multimodal fusion.
We selected three input modalities (RGB, NDVI, and NIR) for multimodal RTS mapping to achieve a robust feature representation. Specifically, the optical image comprises three channels (RGB), while NDVI and NIR each have one channel. To align the input data with the processing requirements of the multi-scale ViT, each modality was rescaled and normalized. Since NDVI and NIR only have one channel, they were duplicated to create three-channel inputs, ensuring compatibility with the multi-scale ViT architecture.
Pre-training singular-modality data independently can fully analyze and extract important image features within each modality, avoiding cross-modal disturbances \cite{wang2021geoai}. It can also considerably alleviate memory demands, as will be demonstrated in our following experiments. This approach also sets the stage for subsequent training with flexible combinations of modalities without the need to start the entire process from scratch. Furthermore, it ensures that RTS mapping results, derived from different multimodal combinations, can be compared fairly under a consistent configuration. 

\subsection{Residual cross-modality attention fusion}
Once we obtain a pre-trained model for each modality, we integrate their backbones for multimodal RTS mapping. Each multi-scale ViT backbone processes its designated modality input to generate multi-scale feature maps, denoted as $F_m \in R^{C×H×W}$, where the feature maps $F$ at each scale for each modality $m \in {RGB, NDVI, NIR}$ has dimensions $H × W$ with $C$ channels. When incorporating N modalities, where $N \in {1, 2, 3}$, the combined feature map at each scale has a total of $N × C$ channels. Since Cascade Mask R-CNN is originally designed for single-modality input with $C$ channels, it is necessary to adjust the model to accommodate the increased number of channels for multimodal input.  
Previous studies \cite{ophoff2019exploring, wang2021geoai} have adopted a “single convolutional layer” to reduce the number of channels. Feature maps from multiple modalities are concatenated together along the channel dimension and processed through one convolutional layer to create fused feature maps. However, this method has a limited capacity for feature representation, as it could disrupt the learned knowledge of each modality. Meanwhile, the fused feature maps, without the information of individual data modalities, can not fully unleash their potential for leveraging multimodal data \cite{he2016identity}. As a result, feature-level fusion using a single convolutional layer could lead to a substantial loss of valuable original information. An efficient fusion strategy aims to harness information from all available modalities, enhancing the performance of a singular modality while minimizing information loss and cross-modal interference. Given the large size of the model, which includes at least one backbone branch per modality, the fusion module must be efficient and lightweight. 

To meet these requirements, we developed a feature-level fusion strategy called the "residual cross-modality attention fusion strategy" (refer to Figure \ref{fig_arch} for its visual representation). This strategy not only enables the model to learn new features from multiple modalities through the cross-attention operation but also incorporates each modality’s original feature information into the new fused feature maps through the residual connection fusion strategy. As a result, more critical information is preserved and enriched with our proposed fusion strategy.
During the pre-training stage, we observed that the model using the RGB modality achieved higher prediction accuracy than the other two modalities in a unimodal model. Therefore, our fusion strategy was tailored to enhance the primary RGB modality with either NDVI or NIR as auxiliary resources. This results in two separate fusion processes: one combining NDVI with RGB and the other combining NIR with RGB. In this approach, each supplementary modality (NDVI or NIR) is fused independently without mutual interference. It also provides flexibility for any modality combination that includes RGB without requiring changes to the architecture of the fusion module.
When fusing any two modalities, we propose using a cross-attention operation instead of the conventional convolutional fusion. Unlike a self-attention operation, which derives important parameters Q (query), K (key), and V (value) from a single modality, the cross-attention module uses RGB feature patches as the query and feature patches from the other modality (NDVI or NIR) as the key and value. This setup enables RGB to extract complementary information from auxiliary modalities, reinforcing visible-spectrum features in a controlled manner. The outputs of these cross-modality interactions are further combined with residual components to form a composite representation. 
The design of the residual fusion module is inspired by the seminal convolutional neural network, ResNet \cite{he2016identity}. In this network, a residual connection module connects the input of a convolutional block directly to its output, effectively bypassing intermediate layers. This mechanism enables the integration of important information across layers, boosting model performance and overcoming the vanishing gradient problem commonly observed in deep networks. The concept of residual connections has also been adopted in subsequent works, such as DenseNet \cite{huang2017densely} and ResNeXt \cite{xie2017aggregated}, further demonstrating its effectiveness in improving deep learning models. 

Specifically, when fusing all three modalities in which RGB modality is required, the feature maps of one modality (e.g., NDVI), represented as $F_{mod1}$, and RGB ($F_{rgb}$) are fused through a cross-attention layer, $CrossAttn_1$. The process enables the RGB features to dynamically attend to relevant information in the NDVI features, resulting in a fused feature representation $F'_1$. Similarly, the feature maps of another modality (e.g., NIR), $F_{mod1}$, and RGB, $F_{rgb}$, are fused using another cross-attention fusion layer, $CrossAttn_2$, yielding the fused feature $F'_2$, as shown in the following equations: 

\begin{equation}
    F'_1 = CrossAttn_1(F_{rgb}, F_{mod1})
\end{equation}
\begin{equation}
    F'_2 = CrossAttn_2(F_{rgb}, F_{mod2})
\end{equation}

The original feature maps for RGB ($F_{rgb}$) and another two modalities ($F_{mod1}$ and $F_{mod2}$) are then fused with the previously generated features ($F'_1$ and $F'_2$) through an element-wise addition operation. Finally, a projection layer, $Conv$, is further applied to the result, as shown in the equation:

\begin{equation}
    F^"=Conv(F_{rgb} \oplus F_{mod1} \oplus F_{mod2} \oplus F'_1 \oplus F'_2)
\end{equation}

Where $\oplus$ denotes element-wise addition, and $Conv$ is configured with both input and output channels set to $C$, the default number of channels in Cascade Mask R-CNN. The resulting feature map, $F^"$, serves as the input for subsequent classification and regression tasks.
When the number of input modalities differs from three, the fusion process adapts. Each modality interacts directly with the RGB modality through cross-attention mechanisms. In the case of two modalities, the RGB modality interacts with the other modality via a single cross-attention layer, generating a fused representation. If there are more than three modalities, each additional modality, beyond the RGB, sequentially interacts with the RGB features using its own cross-attention layer. This approach maintains the integrity and dimensions of the feature maps without needing extra layers for specific channel adjustments. Finally, all generated and original feature maps are combined using element-wise addition. A convolutional layer is then applied to this combined result, ensuring the integration of all fused and original features. 
After the feature fusion, the resulting feature maps are sent to the decoder for instance segmentation (see Figure \ref{fig_arch}). Specifically, the region proposal network (RPN) uses feature maps to generate object proposals indicating objects’ locations \cite{hsu2021knowledge}. These proposals, along with the feature maps, are then sent to its multi-stage heads, where each head refines results from the previous stage to enrich feature information. In the end, these refined features are employed by the model’s segmentation head to generate precise pixel-level masks.

\section{Experiements}
We evaluated our proposed method using the RTS dataset derived from our study area. The entire RTS dataset comprises 717 training samples and 138 testing images. Each data sample is stored in a GeoTIFF file, containing RGB, NDVI, and NIR data, along with binary masks serving as ground truth. To convert these binary masks into instance-level ground truth, we used OpenCV's ``findContours" function, which identifies the contours of distinct objects in the mask. We implemented the model based on the Detectron2 library developed by Facebook AI Research \cite{wu2019detectron2}. Detectron2 provides state-of-the-art detection and segmentation algorithms and offers a complete workflow for both training and testing, suitable for research and production applications. For our model, we integrated the proposed multimodal fusion strategy into Cascade Mask R-CNN \cite{cai2019cascade} for instance segmentation. MViTv2 \cite{li2022mvitv2} was used as its feature extraction backbone.

Training was conducted on a GPU server with four Nvidia RTX A5000 GPUs, each equipped with 24 GB of memory. Each experiment ran for 36 epochs (complete cycles through the dataset) using the AdamW optimizer \cite{loshchilov2019decoupled}, selected for its adaptability and regularization benefits. The loss functions were the same as those used in Mask R-CNN, including classification loss (for accurate object labeling), bounding box regression loss (for precise object localization), and mask loss (for detailed per-pixel segmentation). We adjusted anchor sizes to match the typical dimensions of retrogressive thaw slump (RTS) features, allowing the model to better capture objects of interest. In addition, the learning rate was scaled linearly \cite{goyal2017accurate} to maintain stable performance as training batch size was dynamically adjusted. For the data split, the same approach was used as described in \cite{yang2023mapping}, which involved sampling from each study area to create separate training and testing sets to ensure balanced representation across regions. 

The metric employed for evaluating instance segmentation is Average Precision (AP) at an IoU (Intersection over Union) threshold of 0.5, known as AP50. This metric is widely used in instance segmentation tasks to assess a model's ability to accurately detect and localize objects. AP is calculated as the area under the Precision-Recall (P-R) curve, which plots the model's precision and recall at various confidence thresholds. IoU measures the overlap between the predicted mask and the ground truth by dividing the area of intersection by the combined area of both. Specifically, AP50 refers to the Average Precision when the IoU threshold is 0.5. This means a predicted object is considered correct if its IoU with a ground-truth object is greater than 0.5. To compute AP50, precision and recall are evaluated for all predictions, ranked by their confidence scores. Precision is then interpolated to ensure it does not decrease as recall increases, and the area under this interpolated curve is calculated to obtain the AP value. AP50 provides a concise measure of the model's performance at a relatively lenient IoU threshold.

Three groups of experiments were conducted to assess the efficiency of multimodal learning in RTS mapping. To examine the effectiveness of multi-scale ViT enhanced instance segmentation, we compared mapping results across three models: Mask R-CNN with ResNet50-FPN, Mask R-CNN with multi-scale ViT, and Cascade Mask R-CNN with multi-scale ViT. Each model was tested using a single modality input (RGB, NDVI, or NIR). We also assessed models with pre-trained backbones, considering their parameter size and memory consumption. By comparing different modality combinations with various fusion strategies, we not only determined the optimal modality combination that maximized the mapping precision but also indicated that our proposed fusion approach yields the most promising results.

\begin{table}[t!]
\renewcommand{\arraystretch}{1.3}
\caption{Comparison of AP50 (\%) on RGB, NDVI, and NIR modality among various model configurations}
\label{table_exp_modals}
\centering
\begin{tabular}{c c c c c c}
\hline
\multirow{2}{*}{Model No.} &
\multirow{2}{*}{Model Name} & 
\multirow{2}{*}{Backbone} & 
\multicolumn{3}{c}{Modality} \\\cline{4-6}
& & & RGB & NDVI & NIR \\\hline
1 & \makecell{Mask R-CNN} & \makecell{ResNet50-FPN} & 31.29 & 12.78 & 12.30 \\
2 & \makecell{Mask R-CNN} & \makecell{Multi-scale ViT} & 37.94 & 13.57 & 15.57 \\
3 & \makecell{Cascade Mask R-CNN} & \makecell{Multi-scale ViT} & 39.73 & 15.30 & 18.06 \\
\hline
\end{tabular}
\end{table}

\subsection{Model Comparison}
As the two popular feature extraction backbones, transformer and CNNs have their own advantages and disadvantages. For example, transformer models excel with large datasets while CNN tends to perform well on smaller datasets \cite{dosovitskiy2021image}. When it comes to transferability, transformers often generalize better across different tasks, even when the downstream data is only weakly related to the data used for pretraining \cite{zhou2021convnets, naseer2021intriguing}. However, CNNs are generally more efficient, making them a practical choice for real-time applications or when lightweight backbones are needed \cite{goldblum2024battle}. To evaluate the performance of different instance segmentation networks paired with CNN-based (e.g., ResNet50-FPN) and transformer-based backbones (e.g., ViT and multi-scale ViT) for RTS mapping, we tested three models applied on the RGB, NDVI, and NIR modalities respectively. The comparative models included Mask R-CNN with ResNet50-FPN (Model 1), Mask R-CNN with multi-scale ViT (Model 2), and Cascade Mask R-CNN with multi-scale ViT (Model 3). ResNet50-FPN serves as a standard CNN-based benchmark because of its well-known performance on various tasks, making it a useful point of comparison for other models or architectures. It has 50 layers and uses residual connections to capture representative features. Meanwhile, Feature Pyramid Networks (FPN) combined with ResNet50 further creates multi-level and hierarchical representation of features at various scales. ViT and multi-scale ViT are transformer-based backbones and they are integrated into the instance segmentation pipelines Mask R-CNN and Cascade Mask R-CNN, respectively. Table \ref{table_exp_modals} shows the performance statistics of these models. 

The results indicate that the RGB modality yields the highest mapping accuracy compared to NDVI and NIR across all three model configurations. When using RGB as the input, the Mask R-CNN with a multi-scale ViT backbone (Model 2) achieves a prediction accuracy of 37.94\%, representing an improvement of 6.65\% over the Mask R-CNN with a ResNet50-FPN backbone (Model 1), which has an AP50 of 31.29\%. This performance advantage is also observed when the other two modalities are used as input, with a performance increase from 12.78\% to 13.57\% for NDVI and from 12.30\% to 15.57\% for NIR, respectively. Model 2's better performance compared to Model 1 is attributed to the capabilities of the multi-scale ViT in extracting important feature representations by capturing data dependencies not only across multiple scales but also over long ranges.
A comparison between different instance segmentation networks, Mask R-CNN and Cascade Mask R-CNN, was conducted. Both networks used the multi-scale ViT as the backbone, as it outperforms CNN-based backbones for this specific task. For all data modalities, Cascade Mask R-CNN (Model 3) achieved higher performance than both Models 1 and 2, demonstrating its superiority in model architecture by adopting multi-stage mask refinement to improve segmentation results. Based on these experimental results, our study builds upon and extends Model 3 to enable multimodal learning using the proposed residual cross-modality attention fusion strategy.

\subsection{Model performance with unimodal pretraining and multimodal fine tuning}
\begin{figure}[!t]
\subfloat[Model Parameters]{\includegraphics[width=.32\textwidth]{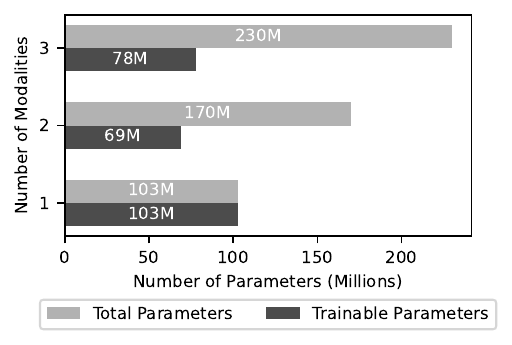}%
\label{fig_exp_mdl_param}}
\hfill
\subfloat[Memory Consumption]{\includegraphics[width=.32\textwidth]{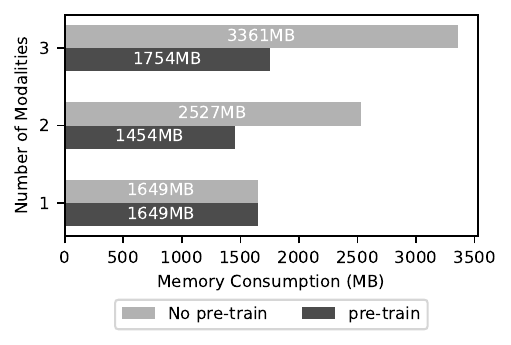}%
\label{fig_exp_mem}}
\hfill
\subfloat[Accuracy]{\includegraphics[width=.32\textwidth]{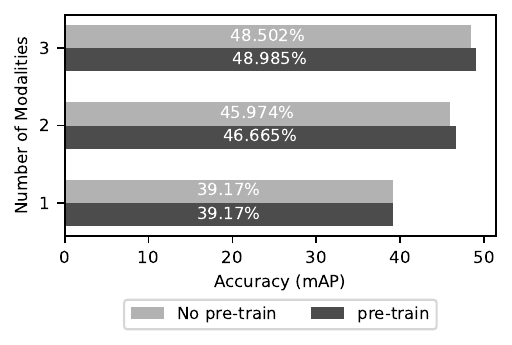}%
\label{fig_exp_accu}}
\hfill
\caption{Statistics on model parameters, memory consumption and accuracy (measured by average precision).}
\label{fig_exp_perf}
\end{figure}

In multimodal training, leveraging the weights of pre-trained backbones without further training has proven to be efficient, as these weights have already achieved the highest AP50 for their respective modalities. Table \ref{table_exp_perf} presents an analysis of the model’s parameters, memory consumption, and accuracy, comparing models with and without pre-trained backbones, with the differences further illustrated in Figure \ref{fig_exp_perf}. For the memory consumption, it is estimated based on the peak memory usage observed during training. This includes the memory required for loading one data sample, loading the model, performing the forward pass, and executing the backward pass. Other potential sources of memory usage, such as optimizer states (e.g., moment estimates in Adam), temporary buffers, CUDA memory caching, and data augmentation, are excluded from this estimation. This consistent baseline provides an accurate comparison of the computational demands for models with and without pre-trained backbones.

\begin{table}[t!]
\renewcommand{\arraystretch}{1.2}
\caption{Model performance and resource usage for different modalities, detailing parameters, memory, and accuracy with and without pre-training.}
\label{table_exp_perf}
\centering
\begin{tabular}{c c c c c c c}
\hline
\multirow{2}{*}{\makecell{\\Number of\\Modalities}} & 
\multirow{2}{*}{\makecell{Number of\\Total\\Parameters\\(Millions)}} & 
\multirow{2}{*}{\makecell{Number of\\Trainable\\Parameters\\(Millions)}} & 
\multicolumn{2}{c}{Memory (MB)} & 
\multicolumn{2}{c}{Accuracy (AP \%)} \\[1.5ex] \cline{4-7}
& & & \makecell{No\\Pre-training} & \makecell{With\\Pre-training} & \makecell{No\\Pre-training} & \makecell{With\\Pre-training} \\[2ex] \hline
1 & 103 & 103 & \multicolumn{2}{c}{1649} & \multicolumn{2}{c}{39.17} \\
2 & 170 & 69 & 2527 & 1454 & 45.97 & 46.67 \\
3 & 230 & 78 & 3361 & 1754 & 48.50 & 48.99 \\
\hline
\end{tabular}
\end{table}

For instance, segmentation models trained on a single modality, such as RGB, NDVI, or NIR, do not require pre-training. These models consist of approximately 103 million (M) parameters, all of which are trainable during the training process. Consequently, the total number of trainable parameters remains 103M, with a memory usage of approximately 1649MB. In contrast, when two modalities (e.g., RGB and NDVI, or RGB and NIR) are utilized, the inclusion of two pre-trained backbones increases the total parameter count to 170M (Table \ref{table_exp_perf}). However, since these backbones are set to be non-trainable, the number of trainable parameters decreases significantly to 69M. This pre-training strategy also reduces memory consumption, requiring only 1454MB compared to 2527MB if all parameters were trainable. While these differences may seem modest, the actual memory usage during training can increase significantly—up to fivefold—when data augmentation is applied. This is due to the quadratic relationship between transformer memory usage and patch sequence length O(n2), in addition to other memory overheads. In terms of performance, the pre-training strategy proved advantageous, achieving a slightly higher AP score of 46.67\% compared to 45.97\% for the non-pre-trained model.

Similarly, integrating all three data modalities (RGB, NDVI, and NIR) leads to a linear increase in the total number of parameters to 230M, but only 78M are trainable when leveraging pre-trained backbones. This configuration offers a substantial reduction in memory requirements, consuming just 1754MB compared to 3361MB when all parameters are trainable. The memory savings become even more critical in practical training scenarios where factors like data augmentation and increased patch sequence lengths can significantly inflate memory usage. From a performance perspective, incorporating pre-trained backbones once again proves beneficial, achieving an accuracy (AP) of 48.99\%, slightly outperforming the 48.50\% achieved without pre-training. These results highlight the scalability and efficiency of the pretraining approach as the number of modalities increases. They also demonstrate that the proposed training strategy maintains comparable or even slightly better performance than the direct training approach, reinforcing its practical applicability.

\subsection{Effectiveness of multimodal fusion strategies for RTS mapping}
In this experiment, we aimed to evaluate the accuracy of RTS mapping utilizing different multi-modalities and fusion methods. Specifically, we compared the efficacy of our proposed residual cross-modality attention fusion strategy with data-level fusion \cite{yang2023mapping}, feature-level convolutional fusion \cite{ophoff2019exploring, wang2021geoai}, stacked-modality attention fusion \cite{xu2023multimodal}, and cross-modality attention fusion \cite{mohla2020fusatnet, yang2024modality, lu2023review, jaegle2021perceiver}. 

\begin{figure}[!t]
\centering
\subfloat[]{\includegraphics[width=.36\textwidth]{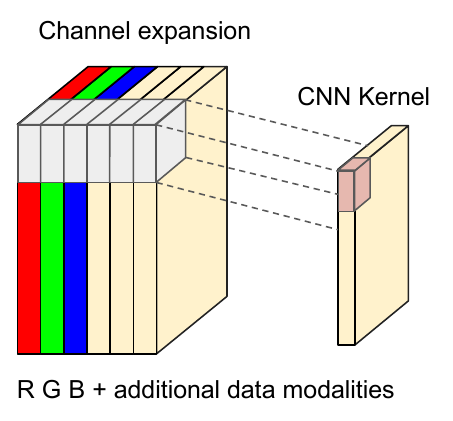}%
\label{fig_exp_fus_chan}}
\hfill
\subfloat[]{\includegraphics[width=.29\textwidth]{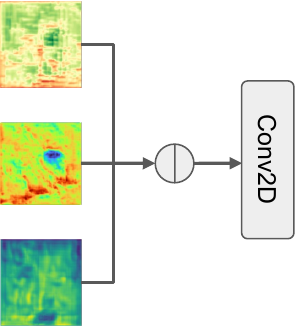}%
\label{fig_exp_fus_feat}}
\hfill
\subfloat[]{\includegraphics[width=.27\textwidth]{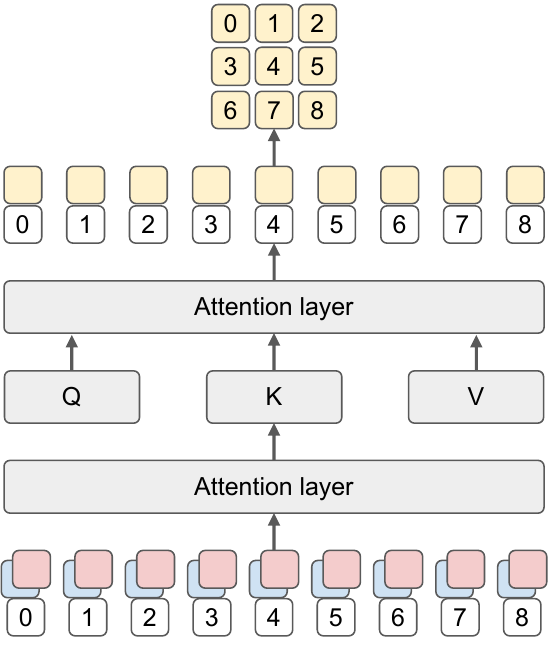}%
\label{fig_exp_fus_attn}}
\caption{Design and implementation of different fusion strategies. (a) Data-level fusion through channel expansion. (b) Feature-level convolutional fusion. Conv2D means a 2D convolutional layer. (c) Stacked-modality attention fusion. Q, K, and V represent the query, key, and value embeddings for attention calculation.}
\label{fig_exp_fuse}
\end{figure}

\begin{enumerate}
    \item data-level fusion: this is the  most commonly adopted approach for incorporating multimodality data, especially in RTS mapping applications \cite{yang2023mapping}. In this approach, different modality data are concatenated along the channel dimension. While simple and easy to implement, this approach requires careful alignment and preprocessing input modalities to ensure consistent spatial resolution. Figure \ref{fig_exp_fus_chan} illustrates the architecture of data-level fusion.

    \item feature-level convolutional fusion: this strategy introduces a convolutional layer that takes the feature maps extracted from each data modality as input, and applies a convolution operation to fuse them into a unified representation. By stacking the feature maps from individual backbones, this method merges complementary information from each modality. Feature-level fusion has been used to detect landform features, such as valleys and mountains \cite{wang2021geoai}. Figure \ref{fig_exp_fus_feat} shows an example architecture design for feature-level convolutional fusion. 
    \item stacked-modality attention fusion: this is a feature-level fusion strategy. The input of the fusion module includes feature maps generated from each modality, and then they are stacked to become a comprehensive feature representation. After that, an attention module is applied to extract feature maps by dynamically analyzing relevant features across the stacked modalities \cite{vaswani2017attention, xu2023multimodal}. Figure \ref{fig_exp_fus_attn} illustrates an example design for stacked-modality attention fusion. 
    \item cross-modality attention fusion: in cross-modal attention fusion, instead of applying the attention module on a single modality, query embeddings from one modality (RGB) will operate with keys and values from other modalities (e.g., NDVI and NIR), enabling conditioned interactions across modalities. This allows the model to focus on relevant features in the context of the RGB data, leveraging the complementary information from other modalities. Unlike stacked-modality attention, where the attention module processes a combined stack of all feature maps, cross-modality attention specifically enhances the understanding of one modality by dynamically attending to the complementary features provided by the others. Cross-modality attention has been effectively used in hyperspectral and lidar classification, as demonstrated by \cite{mohla2020fusatnet} and \cite{yang2024modality}. The subfigure labeled with “Fusion module” in Figure \ref{fig_arch} shows our proposed design and implementation for cross-modality attention fusion. 
\end{enumerate}

Table \ref{table_exp_fuse} provides the model prediction accuracy (measured by AP50) using different fusion strategies and modality combinations. The RGB modality alone is used as a baseline to assess the efficiency of multimodal combinations, given that no fusion strategy is employed in this scenario. The AP50 value of 39.73\% yielded from the singular RGB modality can be regarded as our baseline against which other results can be compared. Employing a data-level fusion strategy with the RGB and NDVI modalities achieves the AP50 of 42.74\%, which surpasses the result from the singular RGB modality. When the NIR modality is added to this combination, the accuracy is further improved to 43.87\%. This result demonstrates the added value of multimodal data in providing complementary information to enhance the model’s predictive capability for delineating RTS. 

\begin{table*}[b!]
\renewcommand{\arraystretch}{1.3}
\caption{AP50 (\%) results for multimodal combinations with various fusion strategies.}
\label{table_exp_fuse}
\centering
\begin{tabular}{c c c c c}
\hline
\multirow{2}{*}{Fusion Type} &
\multirow{2}{*}{Fusion Strategy} &
\multicolumn{3}{c}{Input Modalities} \\\cline{3-5}
& & RGB  & RGB + NDVI  & RGB + NDVI + NIR \\\hline

Data level & Channel expansion  & \multirow{5}{*}{39.73}  & 42.74  & 43.87 \\
\multirow{4}{*}{Feature level} & Convolutional fusion & & 44.51  & 45.45 \\
& Stacked-modality attention fusion & & 44.53  & 45.79 \\
& Cross-modality attention fusion & & 44.71  & 46.72 \\
& \makecell{Residual cross-modality\\attention fusion (our proposed)} & & 46.67  & 48.99 \\[2.4ex]
\hline
\end{tabular}
\end{table*}

Furthermore, all of the comparative feature-level fusion methods outperform data-level fusion across all multimodal combinations. Overall, our proposed residual cross-modality attention fusion approach achieves the highest prediction accuracy of 46.67\% when two modalities (RGB and NDVI) are combined and 48.99\% when three modalities (RGB, NDVI, and NIR) are used in combination (Table \ref{table_exp_fuse}). The advantage of our proposed strategy is attributed to the introduction of both the residual connection module and the cross-modality attention module. The residual connection combines original feature maps from different modalities with fused multimodal feature maps, enabling the model to leverage information from both sources for improved performance. Meanwhile, the cross-modality attention module allows the direct integration of complementary information from other modalities (NDVI and NIR) into the RGB feature space. By integrating features in this way, the model can more effectively capture richer cross-modal dependencies, resulting in superior performance in multimodal learning.

The second-best performing strategy is cross-modality attention fusion (without the residual connection). The convolutional fusion and stacked-modality attention fusion strategies achieve similar levels of performance for both modality combinations, and both outperform data-level fusion. This is because when each modality is processed independently to generate feature maps, the inherent characteristics and unique information of each modality are preserved without cross-modal interference. In contrast, channel expansion in data-level fusion blends features from various modality distributions, potentially introducing noise into the model and degrading performance.

To further evaluate the effectiveness of our proposed fusion strategy, we conducted an ablation study by replacing the cross-modality attention strategy in the residual connection fusion with convolutional fusion and stacked-modality attention fusion. Table \ref{table_exp_ablation} shows the experimental results using different modality inputs. By cross-comparing with the results in Table \ref{table_exp_fuse}, we observe that residual connection fusion, even with the commonly used convolutional fusion to combine multi-modality information, results in better predictive performance (AP50 of 46.21\% when combining RGB and NDVI, and 47.62\% when combining RGB with NDVI and NIR) than using attention-based fusion strategies alone. 

When stacked-modality attention fusion is used without the residual connection module, the model achieves a prediction accuracy of 44.53\% when using RGB and NDVI as input, and 45.79\% when using all three modalities (RGB, NDVI, and NIR) as input (Table \ref{table_exp_fuse}). The cross-modality attention fusion used in the same scenario achieves a prediction accuracy of 44.71\% when using RGB and NDVI as input, and 46.72\% when using all three modalities (RGB, NDVI, and NIR) as input (Table \ref{table_exp_fuse}). Both approaches underperformed compared to residual convolutional fusion (Table \ref{table_exp_ablation}). This finding highlights the effectiveness of residual connection fusion in multimodal learning. Meanwhile, after integrating stacked-modality attention fusion and cross-modality attention fusion, the model performance is further enhanced compared to residual connection fusion, reflecting the effectiveness of attention fusion strategies in multimodal information extraction (Table \ref{table_exp_ablation}). In particular, our proposed residual cross-modality attention fusion achieves the best overall predictive performance at 48.99\% when three modality data are used, demonstrating its capability to fully integrate complementary information from multiple modality data.

\begin{table}[b!]
\renewcommand{\arraystretch}{1.3}
\caption{AP50 (\%) results for ablation study}
\label{table_exp_ablation}
\centering
\begin{tabular}{c c c c}
\hline
\multirow{2}{*}{Fusion Strategy} &
\multicolumn{3}{c}{Input Modalities} \\\cline{2-4}
& RGB  & RGB + NDVI  & \makecell{RGB + NDVI + NIR} \\\hline

\makecell{Residual convolutional fusion} & \multirow{3}{*}{39.73} & 46.21 & 47.62 \\
\makecell{Residual stacked-modality attention fusion} & & 47.00 & 47.69 \\
\makecell{Residual cross-modality attention Fusion} & & 46.67 & 48.99 \\
\hline
\end{tabular}
\end{table}

\subsection{Visualization of RTS mapping results}
\begin{figure*}[!t]
\centering
\includegraphics[width=.95\textwidth]{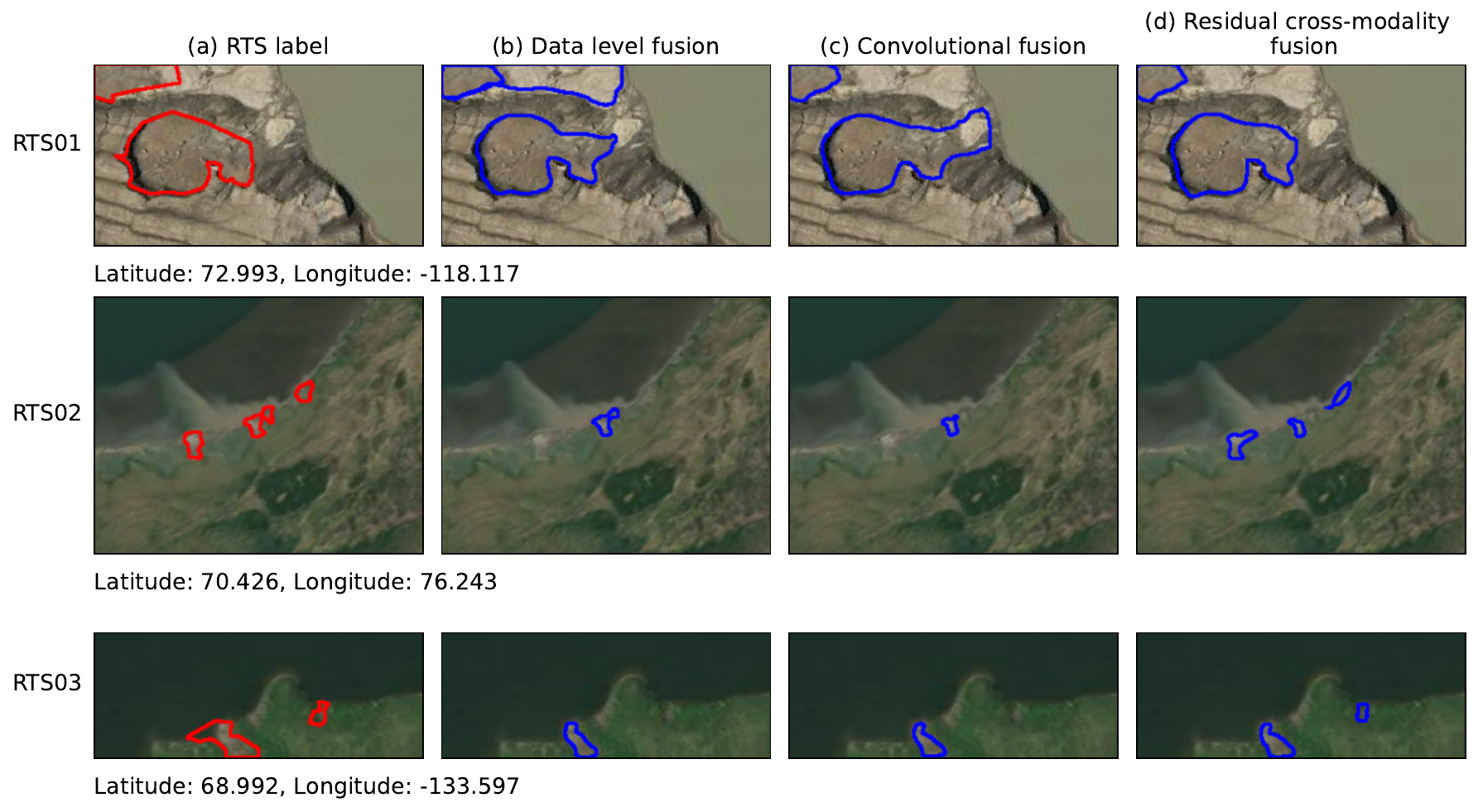}
\caption{Visualization of different model prediction results using combinations of all data modalities.}
\label{fig_vis_pred_all}
\end{figure*}

Figure \ref{fig_vis_pred_all} presents examples of RTS feature predictions. The ground truths are marked in red, and predicted instances are outlined in blue. The first column (a) displays the ground truth instances, while subsequent columns present predictions from one data-level and two feature-level fusion approaches. These approaches utilize all data modalities including RGB, NDVI and NIR. Overall, the results demonstrate that the proposed residual cross-modality fusion outperforms the other two methods. It achieves both recognizing all instances, even when multiple are present, and more precise boundary detection for each instance. 

In the first example, RTS01, all three methods successfully detect the two instances. However, both data level fusion (b) and convolutional fusion (c) mistakenly predict a larger extent for one of the RTS instances. In contrast, the residual cross-modality fusion (d) provides a more precise detection for both instances, closely aligning with the ground truth boundaries. In the second example, RTS02, the ground truth displays multiple instances outlined in red. Both data level fusion and convolutional fusion struggle, detecting only one instance. In contrast, the residual cross-modality fusion effectively detects all instances. Although its boundary accuracy still needs improvement, it demonstrates a stronger ability to recognize general RTS features. Similarly, in the last example, RTS03, the residual cross-modality fusion identifies all instances while the other two methods only capture one instance. 

In conclusion, from the three examples, all three methods are capable of detecting RTS features when they are larger. However, when the features are smaller, only our approach, the residual cross-modality fusion, consistently succeeds in detection. 

\begin{figure*}[!t]
\centering
\includegraphics[width=.95\textwidth]{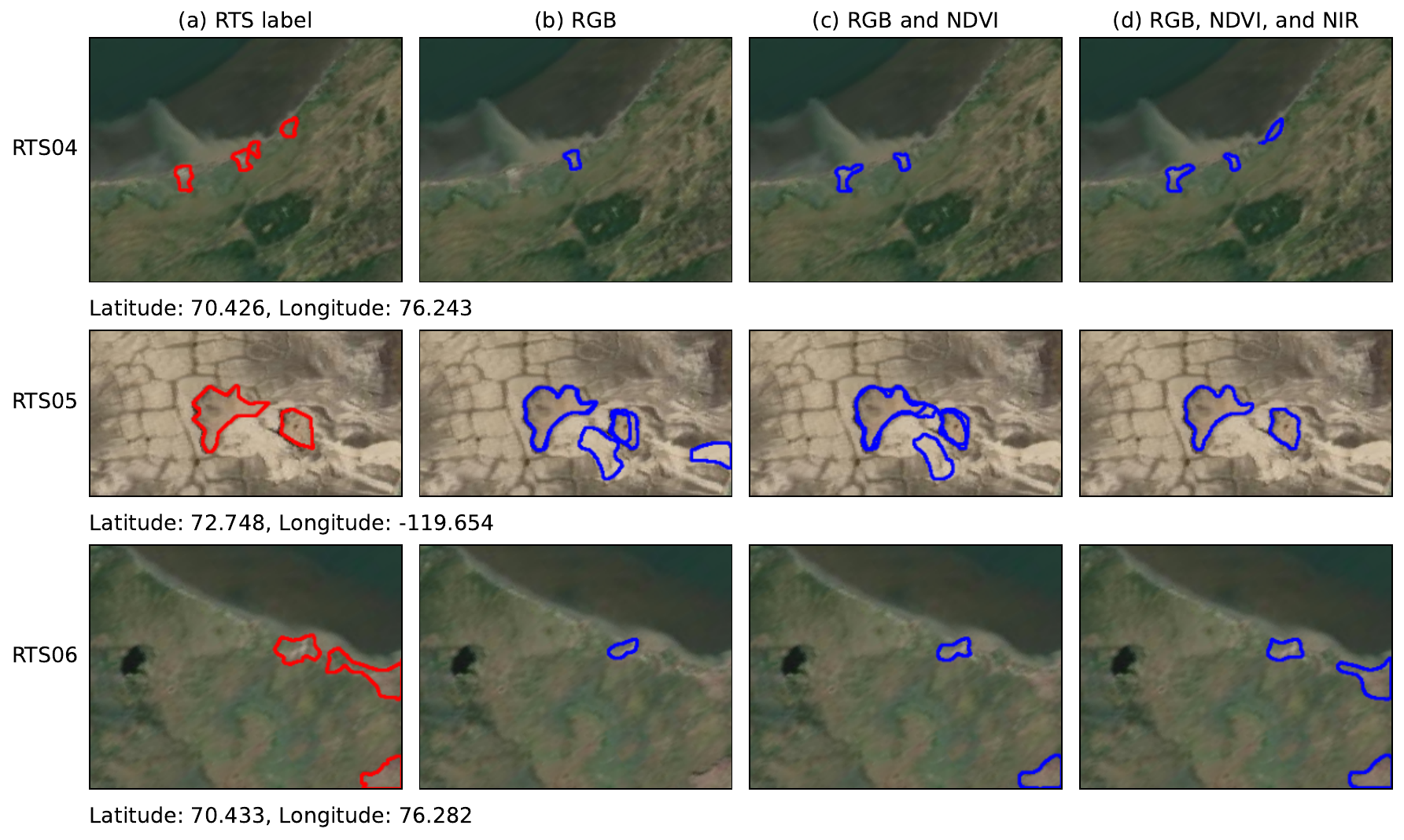}
\caption{Visualization of model prediction results using different data modality combinations.}
\label{fig_vis_pred_modal}
\end{figure*}

Figure \ref{fig_vis_pred_modal} illustrates the impact of various data modality combinations on the detection of RTS features using our proposed method, cross-modality attention fusion. The first column (a) shows the ground truth RTS labels in red. The subsequent columns depict detection results: column (b) uses RGB data, column (c) combines RGB and NDVI, and column (d) integrates RGB, NDVI and NIR. Blue outlines indicate detected features. The progression across columns demonstrates how additional data modalities enhance the model's ability to detect RTS features. 

In the first example, RTS04, using only RGB data provides limited precision. Adding NDVI enhances accuracy and the inclusion of NIR achieves the most precise detection, identifying all three RTS features as the ground truth. Comparing this to Figure \ref{fig_vis_pred_all}, RTS02, which uses the same sample, illustrates that adding more data modalities and effectively fusing them are equally important for achieving optimal results. 

In this RTS06 example, similar to RTS04, the aim is to identify all instances accurately. The results for RTS06 are consistent with RTS04, where integrating more data modalities leads to improved detection accuracy. 

In the RTS05 example, the initial use of RTS data results in duplicate detections, where multiple outlines overlap the same features. It also identifies additional instances compared to the ground truth. Adding NDVI reduces the extra detection but still shows some duplications. The integration of all modalities further refines the detection, effectively eliminating duplicates, extras and aligning closely with the ground truth.  

Overall, the figure demonstrates that incorporating more data modalities improves the RTS detection. As additional modalities like NDVI and NIR are included, the model shows enhanced capability in identifying all instances with greater accuracy. This results in more precise boundary delineations and reduces the occurrence of duplicate instances, highlighting the effectiveness of the cross-modality attention fusion method. 

\subsection{Model generalizability test}
To verify the effectiveness of our proposed multimodal feature fusion strategy, we extended the evaluation on another multimodal landform benchmark dataset, the GeoImageNet \cite{li2023geoimagenet}. GeoImageNet is a natural feature dataset that contains six types of terrain features, including basin, bay, island, lake, ridge, and valley, in both hilly and flat terrains. The training images contain data from two modalities, 1) optical images with 3 RGB bands from 1-meter resolution National Agriculture Imagery Program (NAIP) aerial imagery, and 2) an enhanced DEM data derived from 1/3 arc-second (approximately 10-meter) resolution USGS 3D Elevation Program (3DEP) data. The original DEM data with numerical values are enhanced by compositing multiple rendered DEM derivatives (including color relief, slope, and hill shade) to better align with deep learning paradigms. The features are randomly sampled from the GNIS database, covering 44 states of continental US and Hawaii. There are a total of 876 images labeled by experts through visual inspection, referencing both the USGS Historical Topographic Map Collection (HTMC) data set and NAIP imagery.

\begin{figure*}[!b]
\centering
\subfloat[Basin]{\includegraphics[width=.32\textwidth]{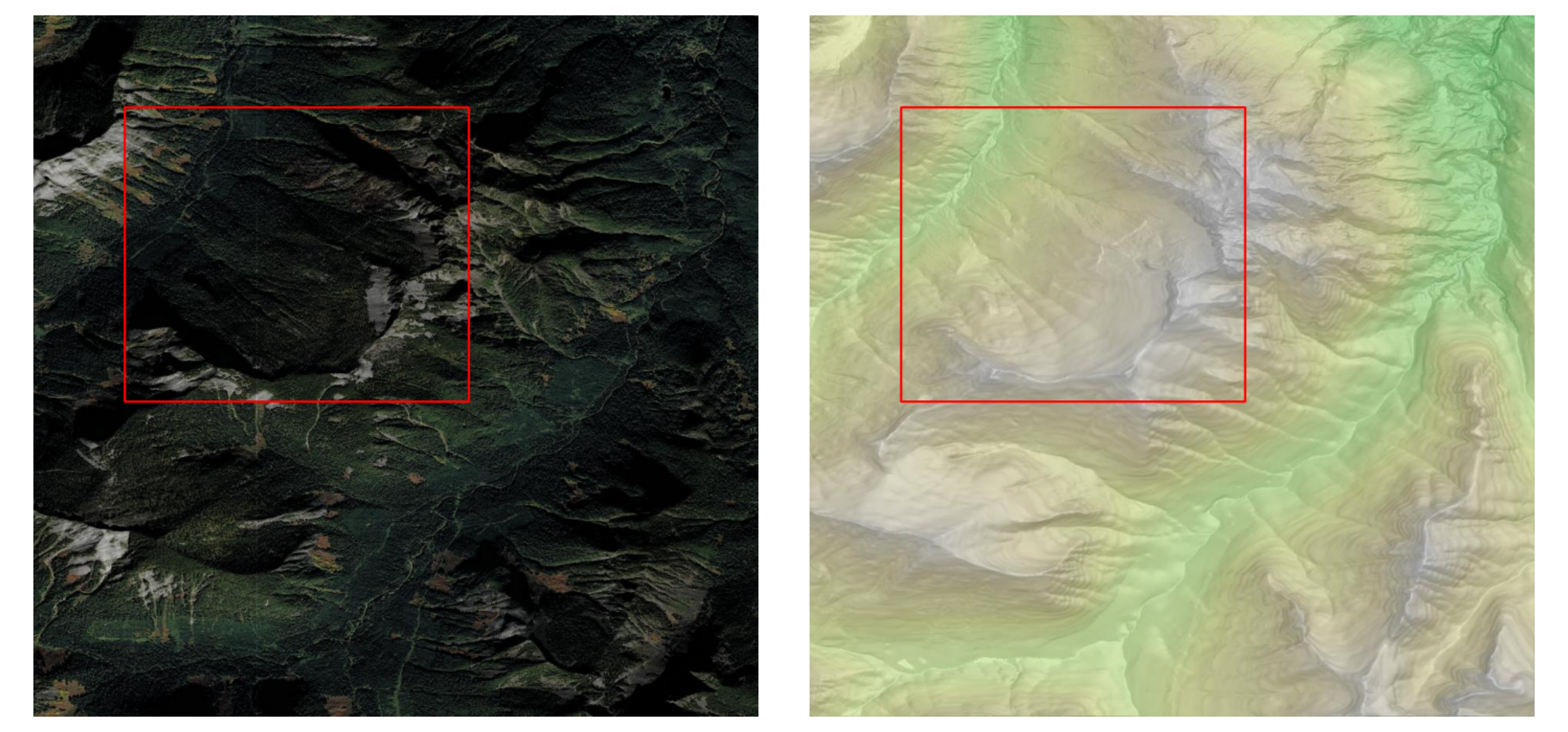}%
\label{fig_vis_catg_basin}}
\subfloat[Bay]{\includegraphics[width=.32\textwidth]{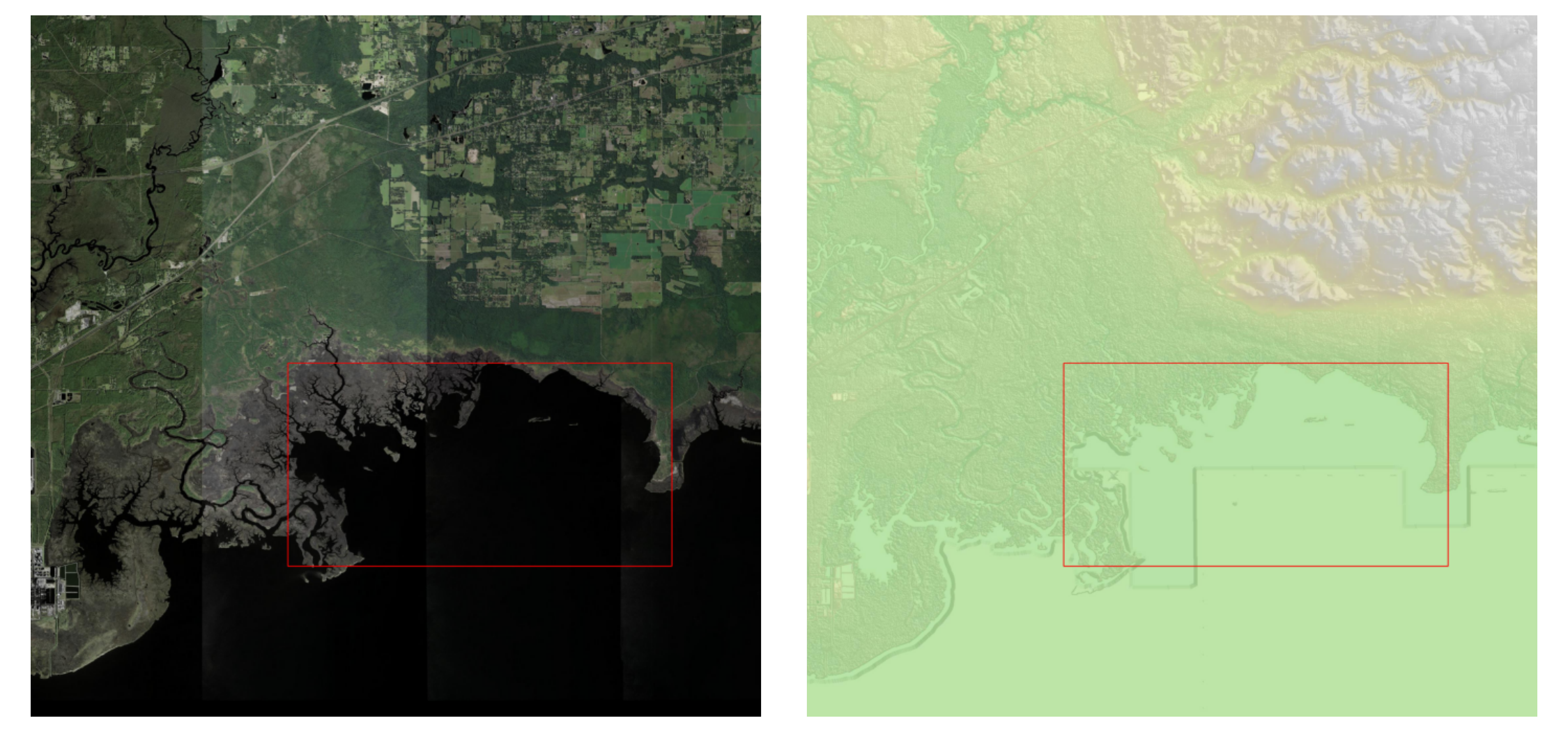}%
\label{fig_vis_catg_bay}}
\subfloat[Island]{\includegraphics[width=.32\textwidth]{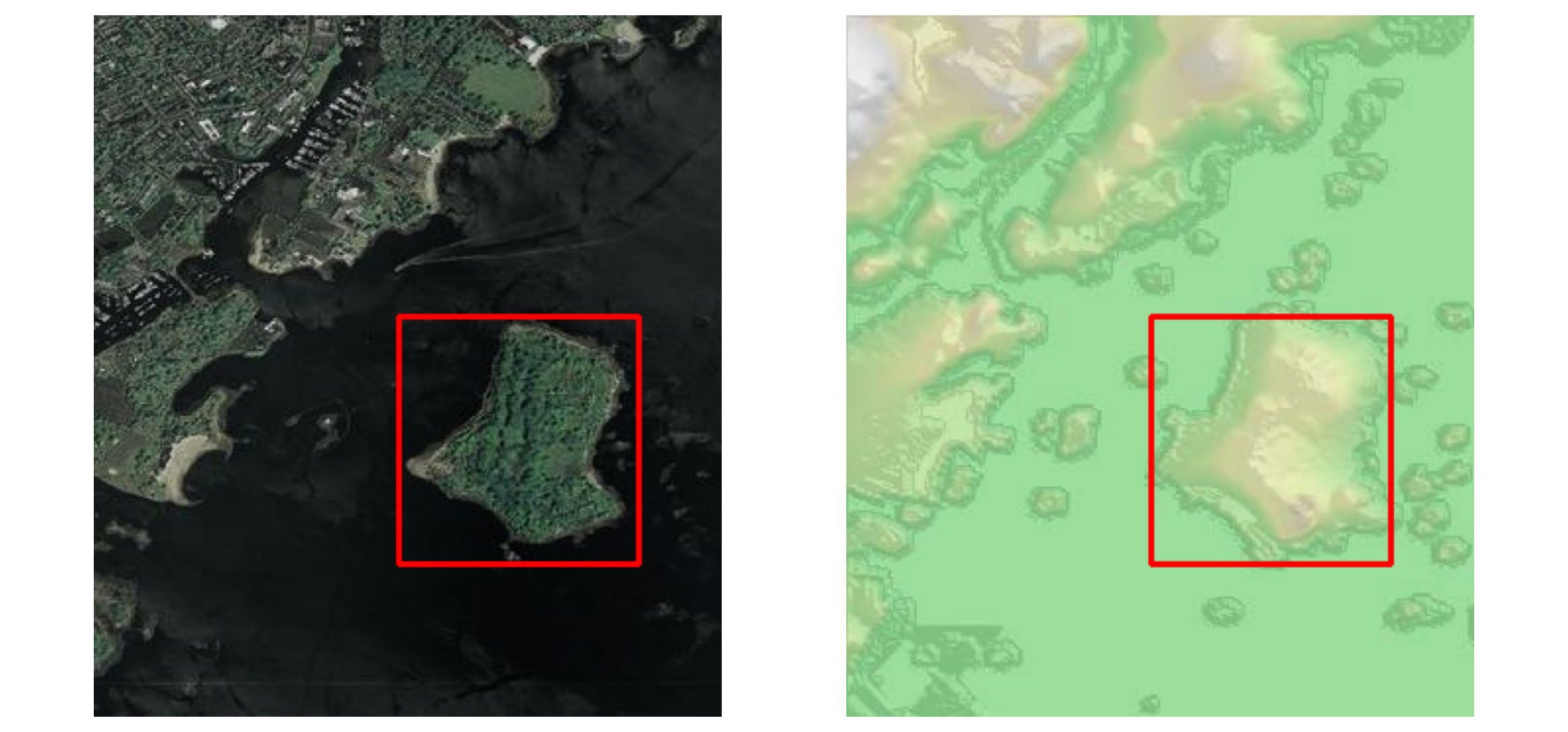}%
\label{fig_vis_catg_island}}
\hfill
\subfloat[Lake]{\includegraphics[width=.32\textwidth]{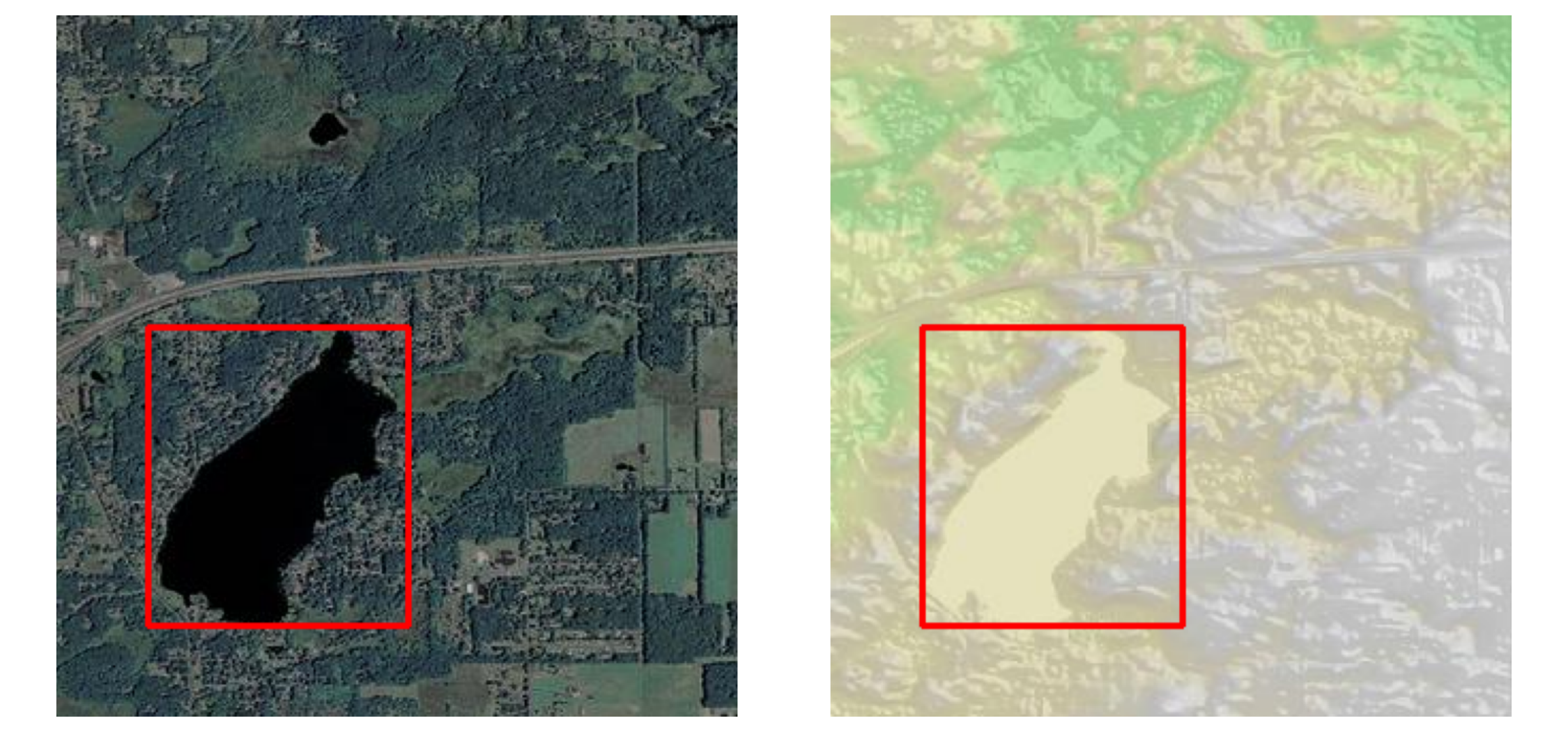}%
\label{fig_vis_catg_lake}}
\subfloat[Ridge]{\includegraphics[width=.32\textwidth]{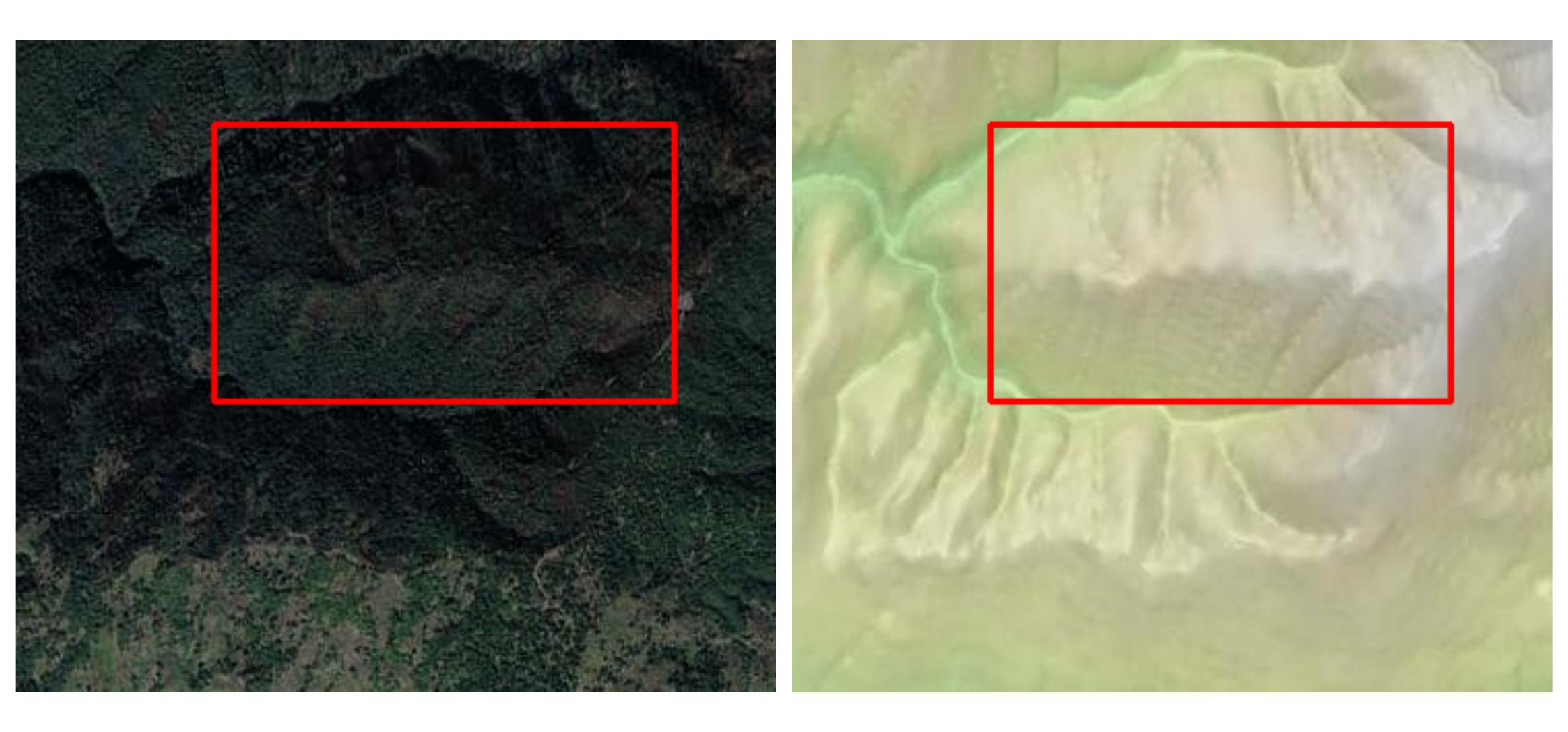}%
\label{fig_vis_catg_ridge}}
\subfloat[Valley]{\includegraphics[width=.32\textwidth]{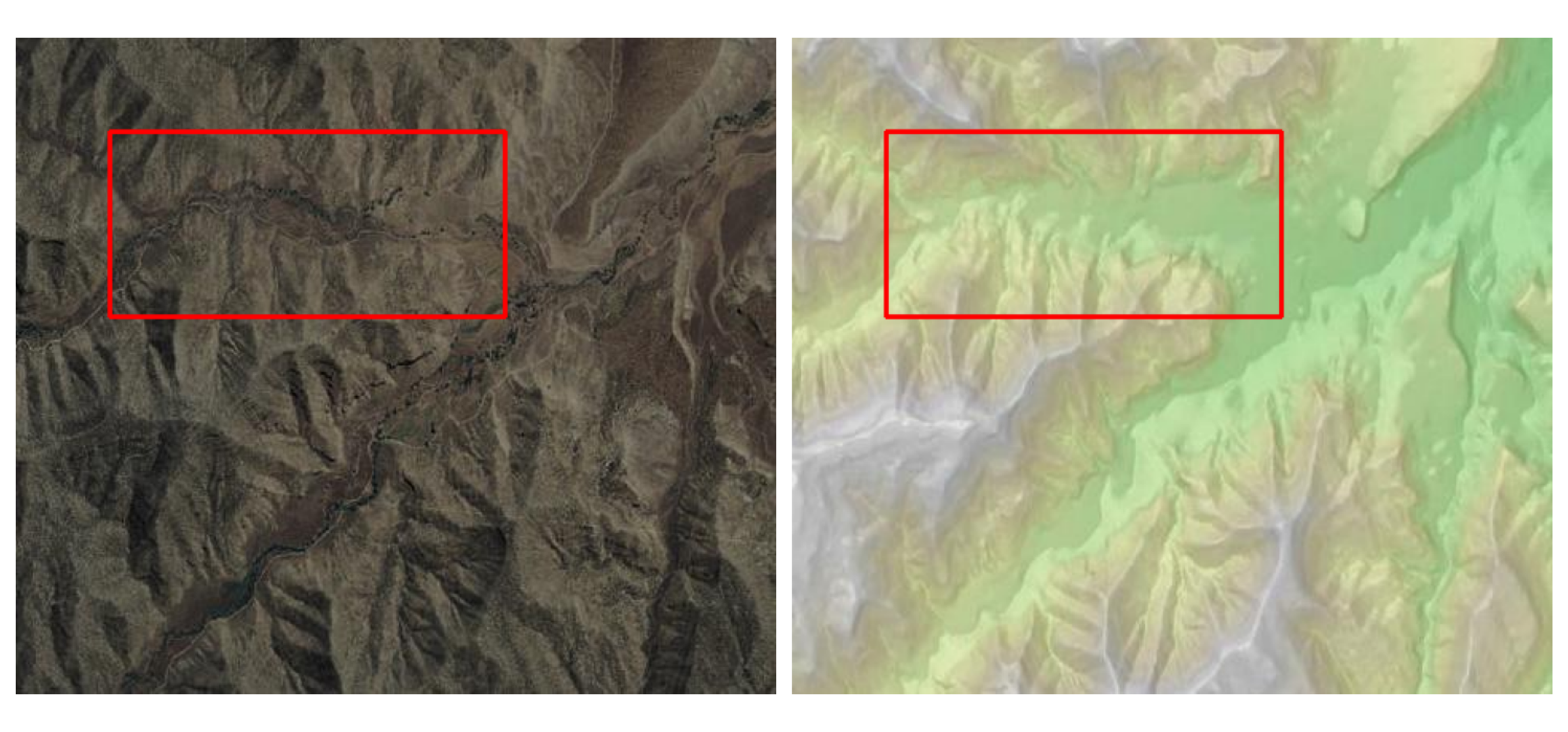}%
\label{fig_vis_catg_valley}}
\caption{Sample training images with their annotations from GeoImageNet.}
\label{fig_vis_catg}
\end{figure*}

Similar to thaw slump dataset, GeoImageNet is also an AI-ready dataset for landform mapping. Figure \ref{fig_vis_catg} shows some example training images with its annotations. Different from the thaw slump dataset, the GeoImageNet data is labeled to support object detection tasks, which include object localization by predicting a bounding box and classification of object class. This variation allows us to evaluate our model's performance and robustness not only among different datasets but also different image analysis tasks. Fortunately, the instance segmentation framework we adopted can easily be adapted to support object detection, by using a different head (decoder). So almost the entire image analysis pipeline, including the feature fusion module, illustrated in Figure \ref{fig_arch} can be reused. The only difference is the removal of the mask head from the Cascade Mask R-CNN architecture. The object detection head is retained to directly predict bounding boxes and class labels. 

\begin{table}
\renewcommand{\arraystretch}{1.3}
\caption{AP50 (\%) results for ablation study}
\label{table_exp_geoimagenet}
\centering
\begin{tabular}{c c c c}
\hline
\multirow{2}{*}{Fusion Type} &
\multirow{2}{*}{Fusion Strategy} &
\multicolumn{2}{c}{Input Modalities} \\\cline{3-4}
& & RGB  & RGB + DEM \\\hline

Data level & Channel Expansion & \multirow{5}{*}{73.39} & 82.88 \\
\multirow{4}{*}{\makecell{Feature level}} & Convolutional Fusion & & 85.28 \\
& Residual Convolutional Fusion & & 85.78 \\
& Residual Stacked-Attention Fusion & & 86.50 \\
& Residual Cross-Attention Fusion & & 86.68 \\
\hline
\end{tabular}
\end{table}

Table \ref{table_exp_geoimagenet} shows the model prediction performance using different fusion strategies and data modalities. It can be observed that fusing RGB data with an additional modality—the enriched DEM—substantially increases the model’s predictive accuracy. It is evident that feature-level fusion strategies are more effective than data-level fusion achieved through channel expansion. Specifically, residual convolutional fusion, which combines the original feature maps with those generated after fusion, results in stronger feature representation and outperforms the commonly used convolutional feature fusion method. Further enhancing this approach with an attention fusion module continues to boost performance. Overall, the residual cross-attention fusion achieves the best predictive performance, with a nearly 14\% increase (from 73.39\% to 86.68\%) in AP compared to single-modal learning results and a 3.8\% increase (from 82.88\% to 86.68\%) compared to the data-level fusion strategy. This finding is consistent with experimental results on the retrogressive thaw slump dataset, providing further evidence of the generalizability and robustness of our proposed feature fusion strategy. It also validates the applicability of our method across varied datasets, geographical contexts (e.g., pan-Arctic vs. the contiguous US), and image analysis tasks (e.g., object detection vs. instance segmentation).

\section{Conclusion}
In this paper, we developed a multimodal approach for mapping retrogressive thaw slumps across the Arctic. Two innovative features were introduced to achieve strong model performance. First, we proposed a residual cross-modality attention fusion strategy that integrates the advantages of (1) a residual connection module, which leverages incremental features important for segmenting thaw slump features by enabling information bypass, and (2) a cross-modality attention layer, which enhances the resultant feature map with complementary information from different modalities through the cross-attention mechanism. Second, to address the high memory consumption introduced by multimodality processing and fusion, we introduced a training strategy that adopts unimodality pre-training in Stage 1 and multimodality fine-tuning with a frozen backbone in Stage 2. This approach enables the model to achieve comparable performance with significantly lower memory demands than regular multimodal training. This training strategy also allows the model to be easily expanded to incorporate additional modalities without the need for re-pretraining, ensuring high scalability for multimodal models and improved computing efficiency in terms of resource consumption. In today’s era of large AI models, developing efficient yet powerful models is critical for addressing geospatial challenges, such as permafrost thaw mapping, while preserving AI’s environmental friendliness \cite{li2024geoai}. The superior performance of our proposed model and multimodal fusion strategy is validated not only on the thaw slump datasets but also through its generalizability, as demonstrated on another landform dataset, GeoImageNet.

In the future, we plan to incorporate additional data modalities into our multimodal permafrost thaw framework to further enhance the model’s robustness across heterogeneous Arctic landscapes. In particular, we will evaluate the effectiveness of the ArcticDEM available from the Polar Geospatial Center to support our mapping work. DEM data captures elevation changes and provides the vertical profile of terrain information, making it highly useful for mapping retrogressive thaw slumps, which are formed by ground subsidence and often involve the collapse of land from its surroundings. However, because thaw slumps can be small in scale with minor vertical drop, the DEM data must possess high vertical accuracy in addition to the desired spatial resolution to support high-accuracy RTS mapping.
In addition to continuing multimodal modeling, we also plan to operationalize thaw slump mapping at a pan-Arctic scale. Such a high-resolution dataset is critical for understanding permafrost thaw and for developing new AI-based models to better identify its triggering factors and conduct near-term forecasts of abrupt permafrost thaw. This research, which integrates AI with domain science, will further advance scientific inquiry and foster the development of new knowledge to better understand the world’s changing environment.

\section*{Acknowledgment}
This research is supported in part by Google.org’s Impact Challenge for Climate Innovation Program and the National Science Foundation under awards 2120943, 2230034, 2230035.

\bibliographystyle{abbrv}  
\bibliography{bibtex/bib/main}

\end{document}